\documentclass[lettersize,journal]{IEEEtran}
\usepackage{amsmath,amsfonts}
\usepackage{algorithmic}
\usepackage{algorithm}
\usepackage{array}
\usepackage[]{subfig}
\usepackage[colorlinks=true, urlcolor=blue]{hyperref}
\usepackage{textcomp}
\usepackage{adjustbox}
\usepackage{stfloats}
\usepackage{url}
\usepackage{verbatim}
\usepackage{graphicx}
\usepackage{makecell}
\usepackage{booktabs}
\usepackage{multirow}
\usepackage{threeparttable}
\usepackage{tikz} 
\usepackage{array}

\newcommand*{\affaddr}[1]{#1}
\newcommand*{\affmark}[1][*]{\textsuperscript{#1}}

\begin{document}

\title{CAST-Phys: Contactless Affective States Through Physiological signals Database}


\author{
Joaquim Comas\affmark[1], Alexander Joel Vera\affmark[1], Xavier Vives\affmark[1], Eleonora De Filippi\affmark[2], Alexandre Pereda\affmark[2], Federico Sukno\affmark[1]\\
\affaddr{\affmark[1] Department of Information and Communication Technologies, Pompeu Fabra University, Barcelona, Spain}\\
\affaddr{\affmark[2] Eurecat Centre Tecnològic, Barcelona, Spain}\\
}



\maketitle

\begin{abstract}

In recent years, affective computing and its applications have become a fast-growing research topic. Despite significant advancements, the lack of affective multi-modal datasets remains a major bottleneck in developing accurate emotion recognition systems. Furthermore, the use of contact-based devices during emotion elicitation often unintentionally influences the emotional experience, reducing or altering the genuine spontaneous emotional response. This limitation highlights the need for methods capable of extracting affective cues from multiple modalities without physical contact, such as remote physiological emotion recognition. To address this, we present the Contactless Affective States Through Physiological Signals Database (CAST-Phys), a novel high-quality dataset explicitly designed for multi-modal remote physiological emotion recognition using facial and physiological cues. The dataset includes diverse physiological signals, such as photoplethysmography (PPG), electrodermal activity (EDA), and respiration rate (RR), alongside high-resolution uncompressed facial video recordings, enabling the potential for remote signal recovery. 
Our analysis highlights the crucial role of physiological signals in realistic scenarios where facial expressions alone may not provide sufficient emotional information. Furthermore, we demonstrate the potential of remote multi-modal emotion recognition by evaluating the impact of individual and fused modalities, showcasing its effectiveness in advancing contactless emotion recognition technologies.

The CAST-Phys dataset is publicly available for research purposes upon request at \href{https://ieee-dataport.org/documents/cast-phys}{\textcolor{blue}{IEEE DataPort}} platform.

\end{abstract}

\begin{IEEEkeywords}
 Multi-modal database, Emotion recognition, Remote Photoplethysmography, Physiological signals

\end{IEEEkeywords}

\section{Introduction}

Emotions are a vital part of human life, deeply influencing decision-making, behavior, and social interactions. They guide our actions and shape how we connect with others, playing a key role in every aspect of our interactions and decisions. Recently, Affective Computing  \cite{picard2000affective}, the field focused on developing systems to recognize, interpret, and simulate human emotions, has gained significant attention due to its broad range of applications in areas such as education \cite{shen2009affective, mejbri2022trends, rodriguez2021affective} or healthcare \cite{Liu, altameem2020facial}.

Emotion states are commonly inferred in the literature using different modalities, such as facial expressions, speech, body gestures, and physiological signals. While facial expressions are increasingly popular due to their intuitive nature \cite{tian2001recognizing, bargal2016emotion, chen2016facial}, physiological signals, also known as peripheral signals, offer distinct advantages. These include their growing availability, thanks to the rise of wearable devices, robustness to external visual noise (e.g., illumination), and high fidelity, as physiological reactions are involuntary and difficult to manipulate or conceal, making them more reliable indicators of the underlying emotional states. Several studies have established a significant link between emotional states and the Autonomic Nervous System (ANS) \cite{kreibig2010autonomic, levenson2014autonomic}. The Sympathetic Nervous System (SNS) and the Parasympathetic Nervous System (PNS), both components of the ANS, play a crucial role in regulating various physiological signals. These signals include electrocardiogram (ECG), blood-volume pulse (BVP), electroencephalogram (EEG), electrodermal activity (EDA), breath rhythm (BR), among others. 

While individual modalities such as facial expressions, speech, or physiological signals have been used to infer emotional states, integrating models that combine features from multiple modalities has proven to be more effective. These multi-modal approaches enhance emotion recognition accuracy and have shown promising results in both research contexts and practical applications \cite{soleymani2011multimodal,jung2019utilizing, comas2020end}. To develop these affective models, several emotion recognition databases for human-computer interaction (HCI) have been proposed, often focusing on facial expressions but less frequently combining them with physiological data. Early multimodal databases, such as MAHNOB-HCI \cite{soleymani2011multimodal}, used implicit affective tagging to label content based on emotional responses. The DEAP database \cite{koelstra2011deap} examined emotional brain activity alongside peripheral signals, while the DECAF database \cite{abadi2015decaf} used magnetoencephalography (MEG) sensors to decode physiological responses to affective content. Recent databases like ASCERTAIN \cite{subramanian2016ascertain} and AMIGOS \cite{miranda2018amigos} include visual and physiological data (EEG, ECG, EDA), focusing primarily on peripheral signals. Despite these advancements, none are designed for remote data acquisition through facial video. The UBFC-Phys \cite{sabour2021ubfc} and MMSE \cite{zhang2016multimodal} datasets are the only ones aimed at remote physiological extraction. However, UBFC-Phys focuses only on stress, limiting its emotional scope, while the MMSE is not publicly available. 

While wearable devices for extracting physiological data for emotion recognition are gaining popularity, the placement of sensors on participants can interfere with their emotional experience and may cause discomfort during extended data acquisition sessions. Moreover, recording new benchmarks for affective computing is often costly and complicated by the need to synchronize devices operating at different sampling rates. Additionally, the real-world applicability of emotion recognition through physiological signals is constrained to a limited range of wearable sensors, such as smartwatches and smart rings. To address these challenges, an ideal solution would involve extracting emotional features from a single modality (i.e. facial expressions) and leveraging it to infer additional modalities. This approach could provide supplementary and valuable emotional information without the need for multiple devices, enhancing both feasibility and accuracy. To this end, this work introduces the Contactless Affective States Through Physiological Signals Database (CAST-Phys), a high-quality, multimodal dataset aimed at advancing research in affective state recognition using non-contact physiological signal measurements from facial videos. The dataset comprises recordings from 61 participants exposed to various emotional stimuli. During the experiments, facial uncompressed recordings and multiple physiological signals were captured simultaneously to ensure synchronized data acquisition. During the acquisition physiological signals including BVP, EDA and RR were recorded due to their potential for remote estimation and their strong relationship with the ANS. 
Additionally, we evaluate the impact of individual and fused modalities on emotion recognition using both contact-based and remote methods. Our findings show that non-contact approaches achieve performance comparable to contact-based systems, underscoring their potential for advancing robust and reliable contactless emotion recognition. The contribution of this work is two-fold:

\begin{itemize}
    \item We present the CAST-Phys database, a novel realistic multi-modal emotion recognition dataset designed to explore affective state recognition using both contact and non-contact physiological measurements from facial videos. The dataset is freely available\footnote{https://ieee-dataport.org/documents/cast-phys} for research purposes upon request.  

    \item We analyze the contribution of the proposed baseline approach, which to the best of our knowledge is the first remote multi-modal emotion recognition approach considering three remotely estimated modalities including, facial expressions, remote photoplethysmography (rPPG) and remote respiration signal (rRSP).

    \item We show that, under our experimental settings, physiological signals are more informative than facial expressions.
    
    
\end{itemize}

The remainder of this paper gives details about the data collection, organization, annotation, and analysis of the presented dataset.
\section{Related work}

\subsection{Multi-modal physiological emotion recognition databases}
In recent decades, several emotion recognition datasets have been developed to explore emotion recognition across various modalities. However, only a few have been specifically designed for multi-modal emotion recognition, integrating modalities such as facial expressions, physiological signals, audio, and others. Among these, the MAHNOB-HCI \cite{soleymani2011multimodal} and DEAP \cite{koelstra2011deap} datasets, developed in 2011, were among the first multi-modal datasets. Both datasets recorded facial videos synchronized with physiological signals during emotion elicitation using videos, capturing both discrete and continuous emotional self-assessments. These datasets have become benchmark references for evaluating new approaches and remain widely used nowadays. Following their release, other similar multi-modal datasets emerged, including RECOLA \cite{ringeval2013introducing}, DECAF \cite{abadi2015decaf}, and ASCERTAIN \cite{subramanian2016ascertain}.

Subsequently, more sophisticated datasets were created. In 2016, the Multimodal Spontaneous Emotion Corpus (MMSE) was introduced, later extended to the BP4D+ database. This dataset includes data from 140 participants, featuring synchronized 3D and 2D facial videos, thermal imaging, and physiological data sequences, along with meta-data such as facial features and FACS codes. A subset of this dataset has been successfully used for video-based heart rate estimation \cite{tulyakov2016self, liu2020multi, revanur2021first}, leveraging its photoplethysmography content. However, its potential for remote physiological emotion recognition remains largely unexplored, offering opportunities for future research. 

Researchers such as Miranda et al. \cite{miranda2018amigos} and Song et al. \cite{song2019mped} have also introduced the AMIGOS and MPED datasets, respectively, both of which improve emotion elicitation techniques to evoke more genuine emotional responses while recording physiological signals similar to those in earlier datasets. While all previous datasets were recorded under laboratory conditions, limiting their ecological validity, the DAPPER dataset \cite{shui2021dataset} overcomes this limitation by collecting ambulatory physiological signals through wearable sensors and emotional self-reports from participants during their daily life activities. The dataset presented in this work, CAST-Phys, is intended to facilitate the study of remote emotion recognition through facial analysis, with the long-term goal of enabling affect estimation in real-world scenarios, similar to DAPPER, but without the need for wearable sensors. Finally, the most recent dataset in this field is the UBFC-Phys dataset \cite{sabour2021ubfc}, created in 2021 to study the impact of social stress on both contact and remote physiological responses. Although not specifically an emotion recognition dataset, it is the first dataset designed for remote physiological signal recovery aimed at extracting arousal responses.

Despite significant advancements in creating new emotion recognition benchmarks, none of the previous works were specifically designed to study the recovery of emotions through non-contact physiological signals. To the best of our knowledge, CAST-Phys is the first publicly available database exclusively developed for multi-modal remote physiological emotion recognition.

\subsection{Camera-based physiological signal measurement}

Since Takano et al. \cite{takano2007heart} and Verkruysse et al. \cite{verkruysse2008remote} demonstrated the feasibility of remote HR measurement from facial videos, researchers have developed various methods for physiological data recovery. Techniques include Blind Source Separation \cite{poh2010non, poh2010advancements}, Normalized Least Mean Squares \cite{li2014remote}, self-adaptive matrix completion \cite{tulyakov2016self}, and approaches based on the skin optical reflection model to mitigate motion artifacts \cite{de2013robust, wang2016algorithmic}. More recently, deep learning-based methods \cite{vspetlik2018visual, yu2019remote, perepelkina2020hearttrack, lu2021dual, comas2024deep} have outperformed traditional techniques. Hybrid methods combine traditional approaches with CNNs for feature extraction \cite{niu2019rhythmnet, song2021pulsegan}, while end-to-end models directly predict rPPG signals from facial videos \cite{chen2018deepphys, Yu2019RemotePS}. Transformer-based models \cite{yu2023physformer++, gupta2023radiant, liu2024rppg, comas2025pulseformer} capture long-range spatiotemporal features but face efficiency challenges, whereas lightweight frameworks \cite{liu2023efficientphys, comas2022efficient} focus on computational optimization. Unsupervised methods enhance generalization using meta-learning \cite{lee2020meta, liu2021metaphys}, contrastive learning \cite{gideon2021way, sun2022contrast}, and techniques addressing domain gaps \cite{lu2023neuron, du2023dual, chari2024implicit}. Additionally, data augmentation strategies tackle biases related to motion \cite{paruchuri2024motion, comas2025beatformer}, skin tone \cite{comas2024physflow}, and heart rate distribution \cite{hsieh2022augmentation}.

\subsection{Camera-based physiological emotion recognition}

Recent studies have explored the integration of physiological signals and facial expressions for remote emotion recognition. Benezeth et al. \cite{benezeth2018remote} were among the pioneers in demonstrating the feasibility of assessing emotional states using remote heart rate variability (rHRV). They employed frequency features from rHRV to differentiate between neutral and arousal states. Similarly, other works aimed to recognize emotions through the extraction of remote photoplethysmography (rPPG) signals. Sabour et al. \cite{sabour2019emotional} used remote pulse variability (rPV) to classify emotions into three discrete categories, while Yu et al. \cite{Yu2019RemotePS} applied remote heart rate (HR) analysis on the MAHNOB-HCI database, also extracting rHRV.

Until now, these studies have primarily focused on rHRV to estimate emotional states. Ouzar et al. \cite{ouzar2022video} were the first to propose a bi-modal approach, combining remote HRV with facial features to improve emotion classification. More recently, Li and Peng \cite{li2024end} and Tao et al. \cite{tao2024facial} introduced similar bi-modal methods, combining facial expressions and rPPG signals using attention mechanisms for emotion recognition, applied to the MAHNOB-HCI and DEAP datasets, respectively. Although recent progress has been made, none of the proposed methods employs a multi-modal approach that incorporates more than two modalities, as we investigate in this work. Additionally, most existing methods evaluate their approaches on outdated multi-modal datasets, such as MAHNOB-HCI or DEAP, which were not designed for remote physiological signal sensing. These data sets overlook important factors, such as video compression \cite{mcduff2017impact, nowara2021systematic, comas2024deep} and the use of contact PPG signal acquisition, which are crucial for rPPG recovery. For these reasons, we introduce a novel database for remote physiological emotion recognition that will be presented in the next section. 
\section{Data collection}
\label{sec:Data collection}
\subsection{Participants}
CAST-Phys comprises a total of 1098 videos from 61 participants (5 participants were eliminated due to technical problems or data sharing refusal). The dataset consists of 35 males and 26 females, with ages ranging from 18 to 38 years old, having an age mean of $25.56 \pm 4.58$ years old. 


\subsection{Emotion Elicitation}
\label{Emotion Elicitation}
\begin{figure}[t]
\centering
\includegraphics[width=85mm,height=75mm]{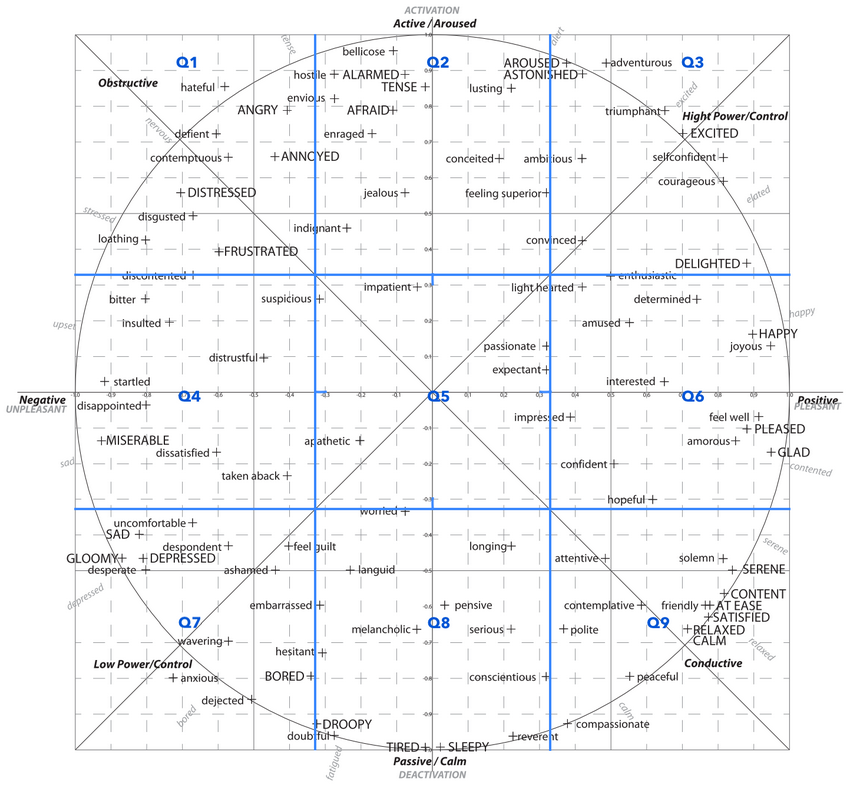}
    \caption{Affective circumplex model divided into 9 regions. The x-axis represents the valence, and the y-axis represents the arousal. Figure is taken from Paltoglou and Thelwall \cite{paltoglou2012seeing}.}
    \label{fig:VA_cimcumplex}
\end{figure}

In this work, the continuous affective representation \cite{russell1980circumplex} was used to characterize the emotional responses elicited by each stimulus. The affect circumplex model is a 2D emotion representation defined by two independent axes: valence and arousal. Valence reflects the level of pleasure or displeasure, while arousal indicates the degree of alertness or activation. To cover the entire valence-arousal domain, we divided the model into nine regions, as shown in Figure \ref{fig:VA_cimcumplex}. 
Based on this segmentation, we have defined two video stimuli that target each of the resulting quadrants, with the objective of creating a dataset that uniformly covers the valence and arousal space, as detailed in the next subsection.

\subsection{Stimuli database}
\label{Stimuli database}

Various methods have been used to elicit emotions, including images \cite{lang1997international}, audio \cite{nardelli2015recognizing}, videos clips \cite{soleymani2011multimodal}, and, more recently, virtual reality (VR) environments \cite{somarathna2022virtual}. For the CAST-Phys dataset, we chose videos as they are multisensory, engaging both visual and auditory senses, without occluding facial regions for remote physiological extraction, unlike VR elicitation. The elicitation dataset consists of 18 videos selected to cover the nine regions of the valence-arousal space defined in Figure \ref{fig:VA_cimcumplex}. 
The videos follow the criteria in \cite{nasoz2004emotion}: short duration (30-60 seconds), self-explanatory visual scenes, and each designed to evoke a single emotion. Additionally, videos with significant spoken content are in Spanish, aligning with the cultural and linguistic context of the participants.

\begin{figure}[b]
\centering
\includegraphics[width=65mm,height=27mm]{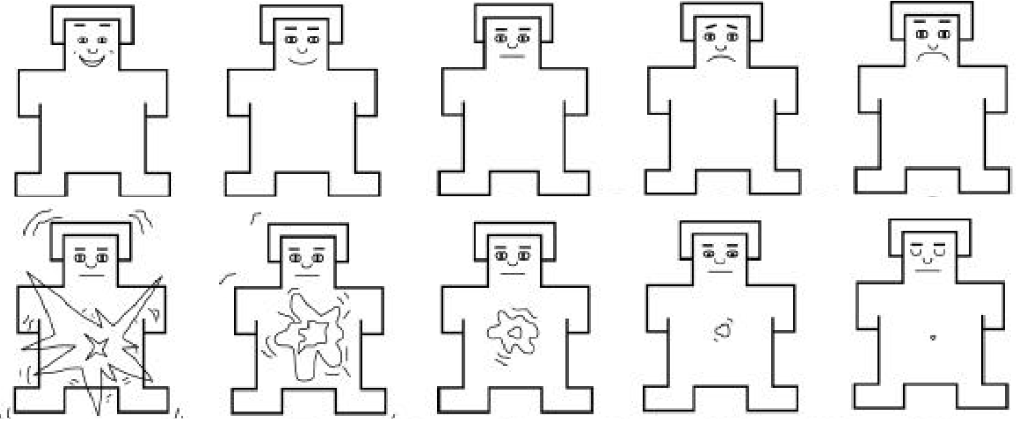}
    \caption{The SAM visual scale used in the study: valence (top row) and arousal (bottom row).}
    \label{fig:sam}
\end{figure}

The stimuli dataset was compiled from various sources \cite{soleymani2011multimodal}, \cite{abadi2015decaf}, \cite{hadar2017implicit}, including publicly available online videos. A total of 56 candidate videos were initially considered to select the final elicitation set. To ensure the selection of the most representative video clips for each emotion, we used a majority-voting strategy with four external annotators. Each annotator independently rated the valence and arousal levels evoked by each video. After completing their annotations, videos that received full agreement among the annotators were included in the elicitation dataset. The selection was conducted in five iterative rounds, where non-consensual videos were progressively replaced by new candidates until full agreement was achieved across all circumplex model regions. This rigorous process ensured the videos selected elicited consistent emotional responses. The final list of elicitation video clips used in the CAST-Phys dataset is presented in the supplementary material.

\subsection{Self-report}
\label{Self-report}

As stated in Subsection \ref{Experimental setup}, immediately after each trial, participants were instructed to use the Self-Assessment Manikin (SAM) test \cite{bradley1994measuring} to self-report the emotions they experienced while watching each video. The SAM test utilizes a set of five images, as shown in Figure \ref{fig:sam}, to represent valence and arousal independently. The images range from $-2$ to $+2$, allowing participants to indicate their perceived emotional valence and level of arousal.

\begin{figure}[b]
\centering
\includegraphics[width=85mm,height=37mm]{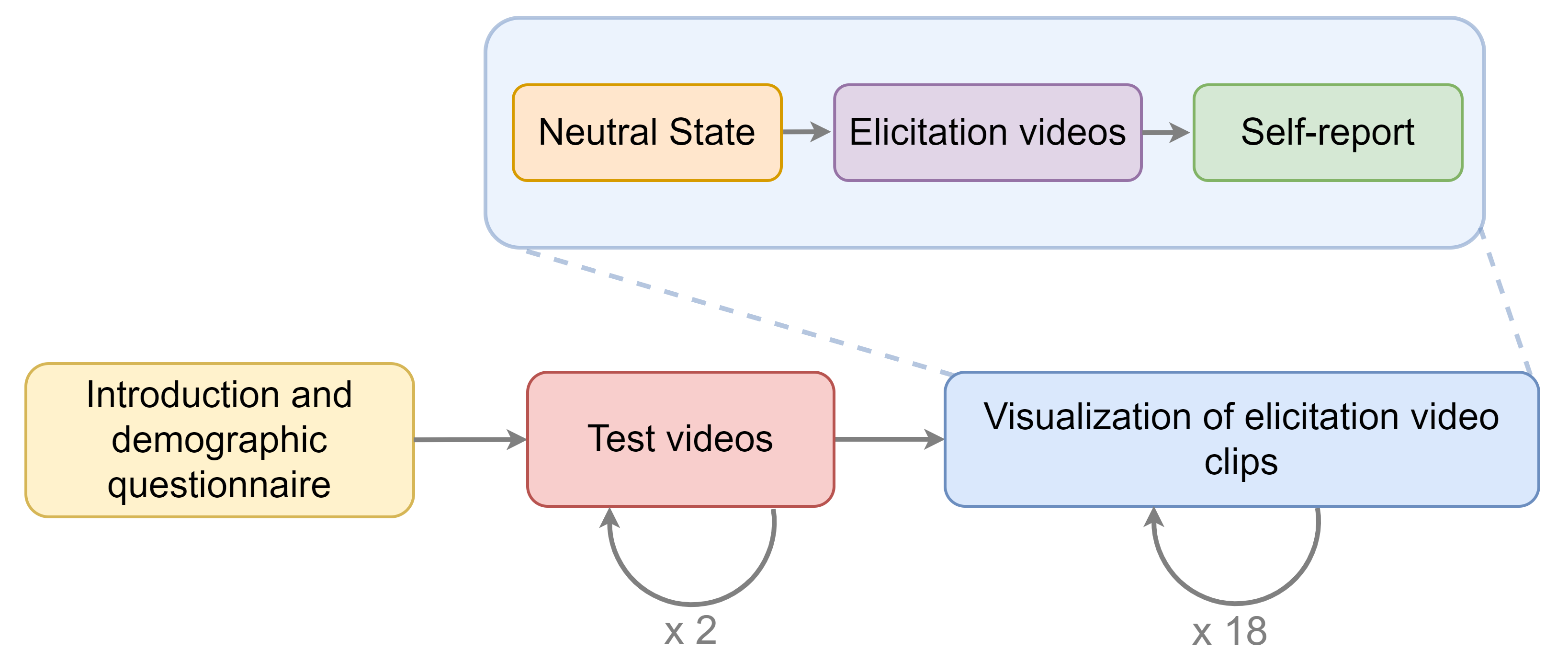}
    \caption{Overview timeline of the experimental sessions.}
    \label{fig:overview}
\end{figure}

\begin{figure}[t]
\centering
\includegraphics[width=85mm,height=57mm]{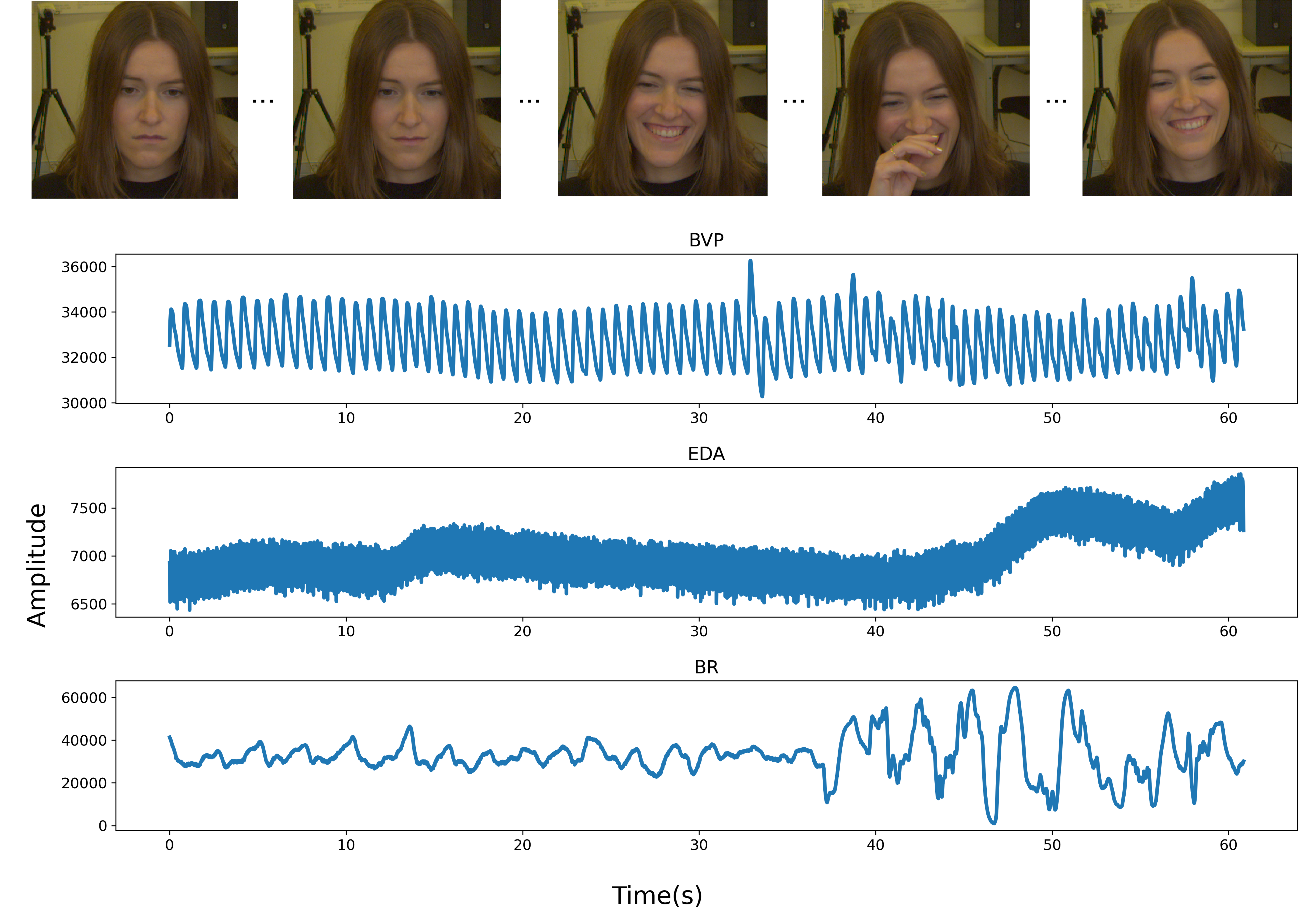}
    \caption{Data collection from a participant during the experiment, including facial recordings and raw physiological signals: BVP, EDA, and BR (top to bottom).}
    \label{fig:modalities}
\end{figure}

\subsection{Experimental protocol and acquisition setup}
\label{Experimental setup}

During the experiment, we first collected demographic information from participants and obtained their consent through the informed consent form, following the protocol approved by the Institutional Committee for Ethical Review of Projects (CIREP) at Pompeu Fabra University, application No. 306. Participants were informed that the experimental results could be published in academic journals or used for research purposes, asking them for permission to share anonymized raw data. 

After providing consent, participants were then introduced to the self-report system and the bio-sensors used in the experiment, followed by an initial test recording. Once data collection began, two test videos were shown, and after each, participants were asked to provide valence and arousal annotations. This stage also allowed participants to ask any questions about the rating system, or the concepts of valence and arousal. The annotations from these two videos were discarded and not used in the analysis. Following this, participants viewed the 18 selected elicitation videos and rated them after each one. A 30-second neutral interval was included between each emotional trial to prevent one emotion from influencing the next. Figure \ref{fig:overview} summarizes the workflow of the experimental sessions.
 
The recording system comprises two devices for synchronized facial video and physiological signal capture. An RGB Basler acA640 camera with a 12mm lens, capable of up to 120 FPS without compression, was configured to record at 60 FPS which was deemed sufficient to avoid disrupting the remote physiological signals. The raw frame resolution was $658\times492$ pixels, which was later cropped and resized to 450x350 pixels centered at the face region to homogenize the final videos and facilitate further processing. Meanwhile, physiological signals were recorded by a Biosignalsplux 4-Channel hub, which supports a maximum sampling rate of 3 kHz. To ensure proper synchronization with the camera, the hub was set to a 60 Hz sampling rate, aligning with the camera's frame rate. The physiological signals were recorded using three bio-sensors: EDA, BVP, and RSP. The EDA sensor, consisting of two electrodes, was directly attached to the skin on the designated fingers. The BVP sensor was placed on the index finger, and the breathing monitor was positioned under the chest. Figure \ref{fig:modalities} shows an example from a participant, displaying the four recorded modalities in a segment of a trial. 

The synchronization among the four modalities is essential for remote physiological signal sensing. To verify the alignment between the facial and physiological data, we computed the temporal offset, which showed an average difference of $8.19 \pm 4.36$ milliseconds for the whole dataset, enabling precise temporal alignment for multimodal data acquisition. 

On the other hand, to allow participants to watch the videos and self-report their emotions, a graphic user interface (GUI) was developed using Psychopy v2023.1.0 \cite{peirce2019psychopy2}. The GUI ran on a separate computer from the data acquisition system, requiring synchronization between both computers to track when the participants were watching the videos. A LED light\footnote{https://thepihut.com/products/tri-color-usb-controlled-tower-light-with-buzzer}, positioned behind the participant, was used to mark the presentation of each stimulus. Controlled by the software running the GUI, the LED blinked a green light when a video started and ended, indicating the timing of the stimulus in the video recording. 

To assess the accuracy of this synchronization, we extracted timestamps corresponding to the first and last LED flashes for every participant and trial. These timestamps were used to compute the effective stimulus duration and compared against the nominal duration of the selected elicitation videos. Across 1098 trial samples, the analysis revealed a Mean Absolute Error (MAE) of $17.10 \pm 27.32$ ms. Given the monitor refresh rate and camera recording frequency of 60 Hz (frame duration $\approx$ 16.67 ms), this deviation confirms that synchronization precision is limited primarily by frame quantization rather than software drift. This demonstrates that the system achieves frame-level accuracy, enabling precise temporal alignment between the stimulus and the recorded videos.

\begin{figure}[t]
\centering
\includegraphics[width=82mm,height=40mm]{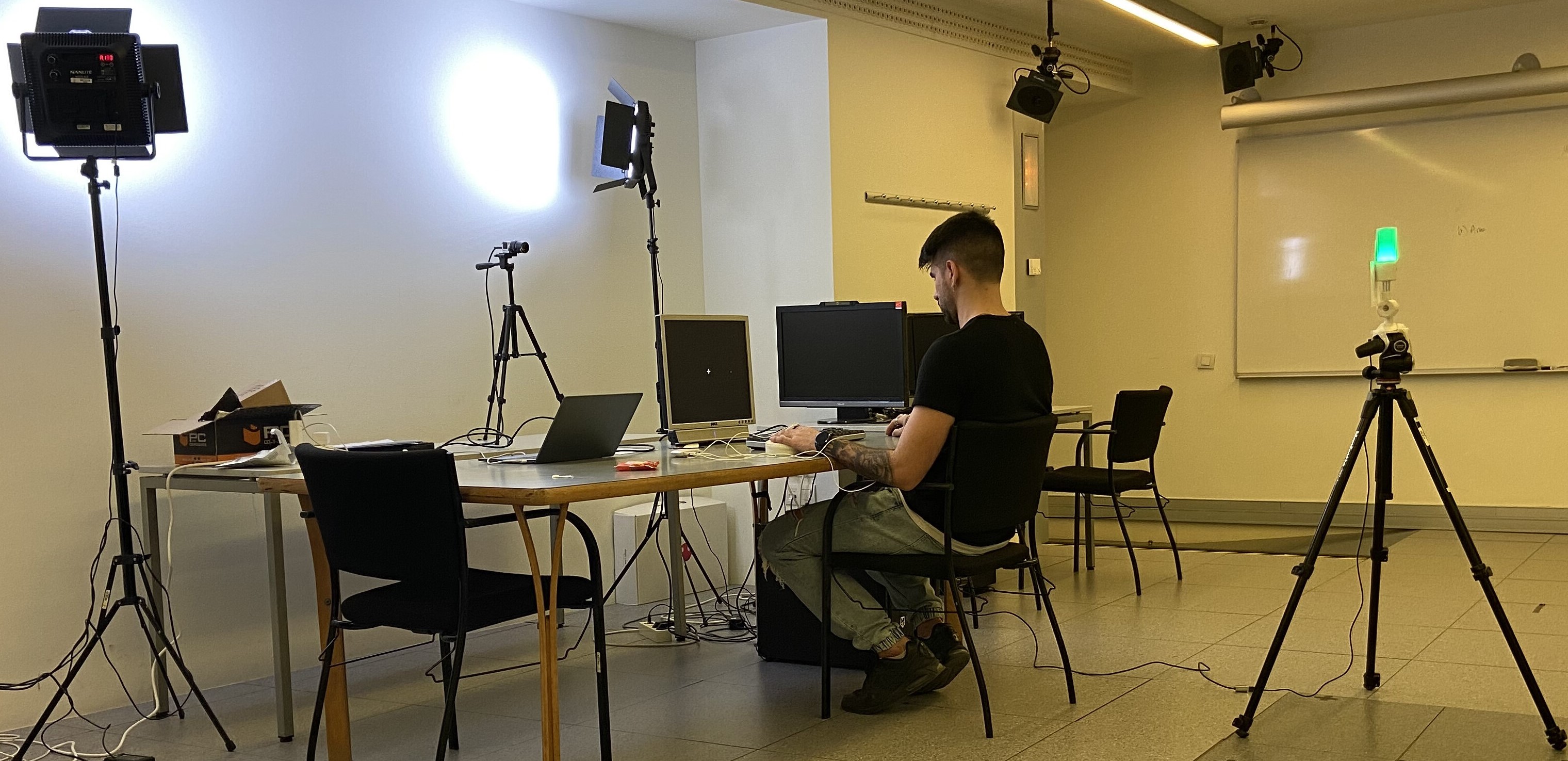}
    \caption{Snapshot of the CAST-Phys experimental setting up.}
    \label{fig:setup}
\end{figure}

Additionally, artificial light sources were used to ensure uniform lighting conditions for all participants. Figure \ref{fig:setup} shows an example of the experimental setup.

\subsection{Database organization}
\label{organization}

The CAST-Phys dataset is organized into 61 folders, each corresponding to a participant. Each participant's folder contains 18 subfolders, labeled according to the nine elicitation regions described in subsection \ref{Emotion Elicitation}. Each trial folder includes the facial video resized and cropped, the emotional self-report of the participant, the physiological signals (BVP, EDA and BR) and the metadata content about the timestamps for each trial. Additionally, each participant folder includes demographic information such as age, gender, and ethnicity, along with metadata detailing the recording duration. The average size of each subject is about 8.5 GB, resulting in over 0.5 TB with about 2.4 million frames in total.

\begin{table}[t]
  \caption{Hand-crafted features extracted from facial and physiological signals.}
  \renewcommand{\arraystretch}{1.35}
  \centering
  \adjustbox{width=0.49\textwidth}{
  \begin{tabular}{ p{2.5cm} p{5cm}}
    \hline
    \hline
    \thead{Modalities} & \thead{Features} \\ 
    \hline
    \hline
    \centering Facial expressions (20 features) & Facial action units: AU01, AU02, AU04, AU05, AU06, AU07, AU09, AU10, AU11, AU12, AU14, AU15, AU17, AU20, AU23, AU24, AU25, AU26, AU28, AU43\\ [0.5ex]
    \hline
    \centering PPG (14 features) & HR mean and standard deviation; HRV: mean and standard deviation of the RR intervals, square root of the mean of squared differences between adjacent RR intervals (RMSSD), standard deviation of the successive differences between RR intervals (SDSD), percentage of absolute differences in successive RR intervals greater than 50 ms (pNN50), baseline width of the RR intervals distribution (TINN), Poincaré plot short-term RR interval fluctuations (SD1), Poincaré plot long-term HRV changes (SD2), ratio of SD1 to SD2 (SD1SD2), spectral power of low frequencies LF (0.04-0.15 Hz), spectral power of high frequencies HF (0.15-0.4 Hz), ratio between LF and HF (LFHF) \\ [0.5ex]
    \hline
    \centering EDA (6 features) & Tonic component mean and standard deviation,  skin conductance response peaks and mean of peaks magnitude, Phasic component mean and standard deviation \\ [0.5ex]
    \hline
    \centering Breathing (4 features) & Respiration Rate: mean and standard deviation, mean of derivative (variation of respiration signal), Respiratory Volume per Time (RVT)  \\ [0.5ex]
  \end{tabular}
  }
  \label{table:fatures_used}
\end{table}

\section{Experimental validation and results}

In this section, we detail the preprocessing and selected features from the collected data and analyze participant self-report annotations, described in Subsections \ref{preprocessing} and \ref{selfreport}. Subsequently, in Subsection \ref{multi_contact}, we present the emotion recognition baseline, evaluating each collected modality's influence and combined performance. Finally, we outline the remote extraction of physiological signals and assess their effectiveness in emotion recognition, as discussed in Subsections \ref{remote_ext} and \ref{multi_remo}, respectively.

\subsection{Data processing}
\label{preprocessing}

The experimental validation was conducted using a total of 1098 trials, corresponding to 18 elicitation videos per participant across 61 participants. Owing to the accurate acquisition and the high quality of the recorded data, no trials were excluded, and all samples were included in the subsequent analyses.

\subsubsection{Physiological signals}

From the three collected physiological signals, various features are extracted using the NeuroKit2 toolbox \cite{Makowski2021neurokit}. Before feature extraction, all physiological signals are standardized by subtracting the mean and dividing by the standard deviation.

For the PPG signal, we derive HR and HRV, which are widely studied due to their strong correlation with emotional responses. From HR and HRV, we extract temporal, spectral, and non-linear features. Temporal features provide insight into beat-to-beat RR intervals, while spectral features, such as the ratio between low and high-frequency bands, offer information about the balance between the SNS and PNS components, also known as sympathovagal balance \cite{Hori1992heart}. Non-linear features, such as the Poincaré plot, reveal correlations between successive RR intervals. In total, we extract 14 features from the HR signal. For the EDA signal, which reflects the SNS activity during high-intensity situations, we mainly considered its two main components: tonic Skin Conductance Level (SCL) and phasic Skin Conductance Response (SCR). SCL represents a smooth baseline level, whereas SCR captures rapid reactions to external stimuli. From the EDA signal, we extract six features.  For the respiration signal, slow breathing is associated with relaxation, whereas irregular rhythms or rapid variations correspond to more aroused emotional states. We extract four respiration features. Across the three physiological modalities, we obtain a total of 24 features.

\subsubsection{Facial Expressions}

From the facial videos recorded during the experiment sessions, we extract facial expression features using the Py-Feat estimator \cite{cheong2023py}. Given the high frame rate of the recordings, we sample facial features at one-second intervals throughout each trial video. For each second of the video, we extract 20 facial features based on the Facial Action Coding System (FACS) \cite{ekman1978facial}, one of the most widely used frameworks for systematically quantifying facial muscle movements. FACS identifies the actions of distinct groups of facial muscles, referred to as action units (AUs), which are recognized as reliable indicators of emotional expression \cite{sayette2001psychometric}.

In total, 44 final features are extracted from physiological and facial modalities, which are summarized in Table \ref{table:fatures_used}. 

\begin{figure}[t]
\centering
\includegraphics[width=85mm,height=77mm]{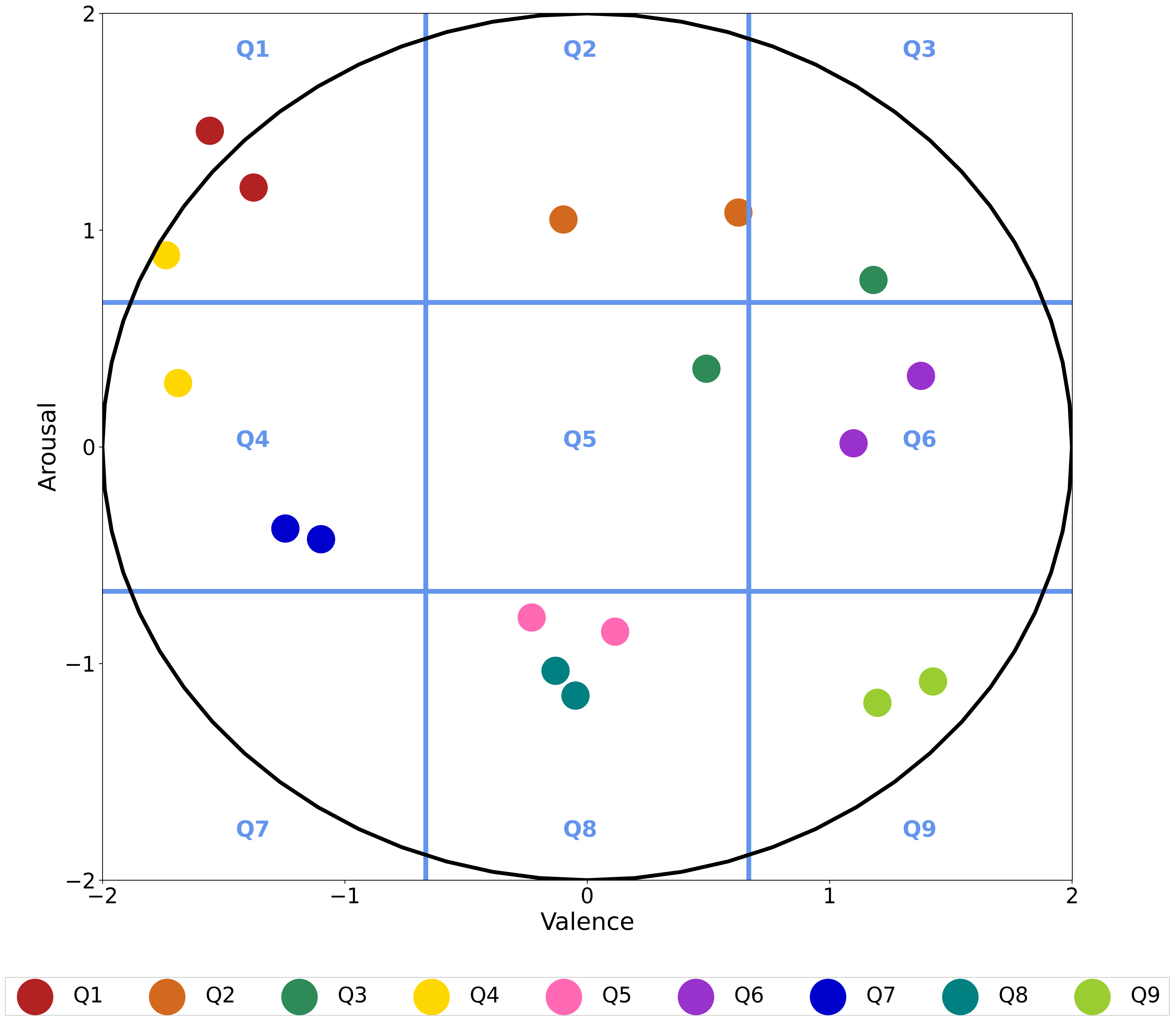}
    \caption{Mean self-report annotations from participants represented in the Affective circumplex model.}
    \label{fig:annotations}
\end{figure}

\begin{figure}[b]
  \centering
  \begin{minipage}[b]{0.31\textwidth}
    \centering
\includegraphics[width=\textwidth]{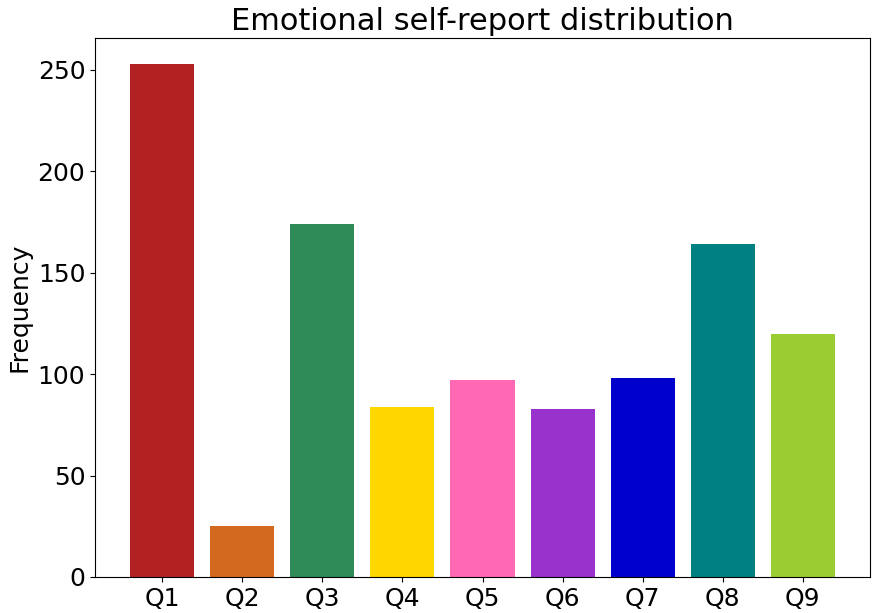}
  \end{minipage}\hfill
  \begin{minipage}[b]{0.16\textwidth}
    \centering
    \subfloat{
\includegraphics[width=\textwidth]{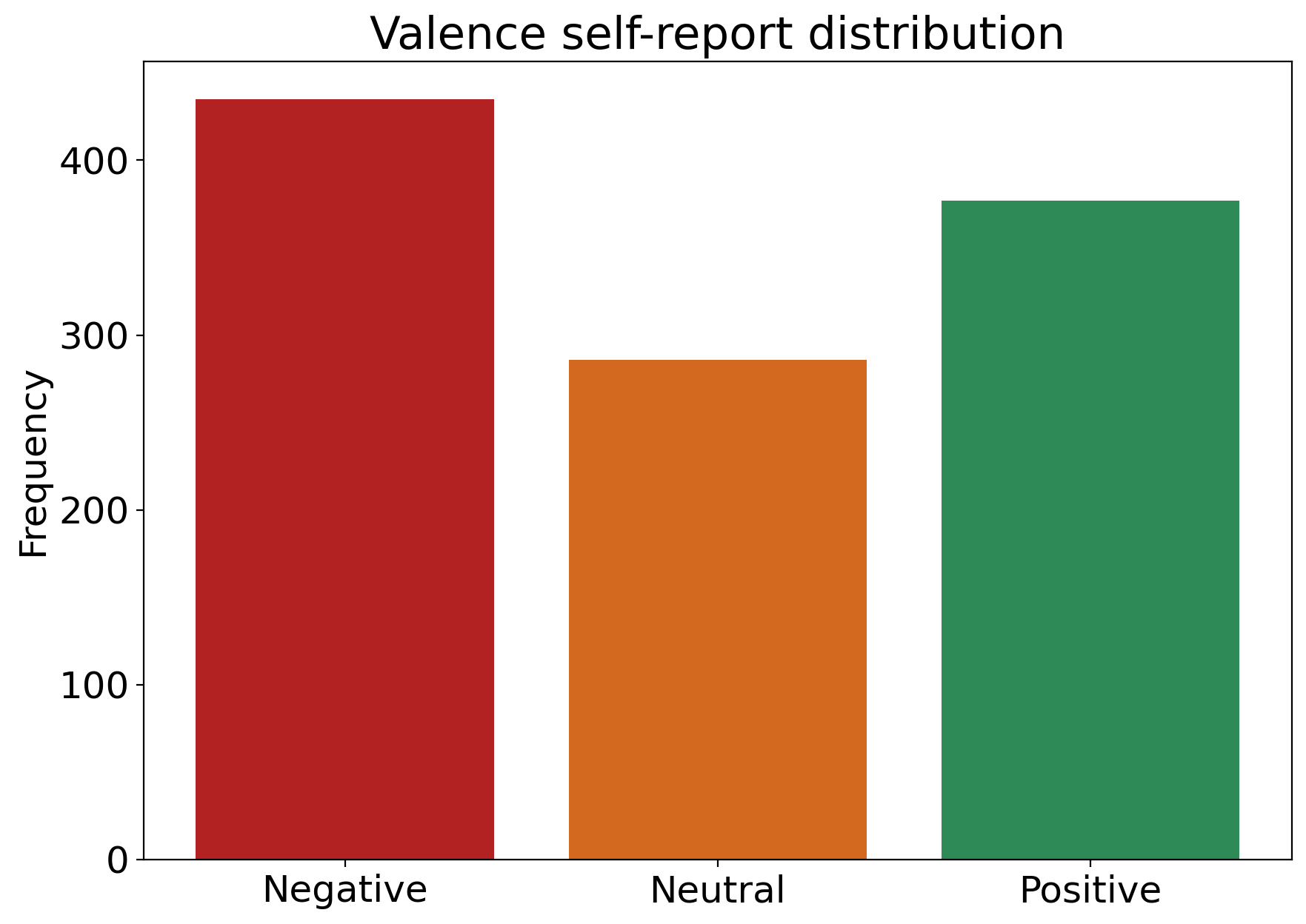}}\\[1ex]
    \subfloat{
\includegraphics[width=\textwidth]{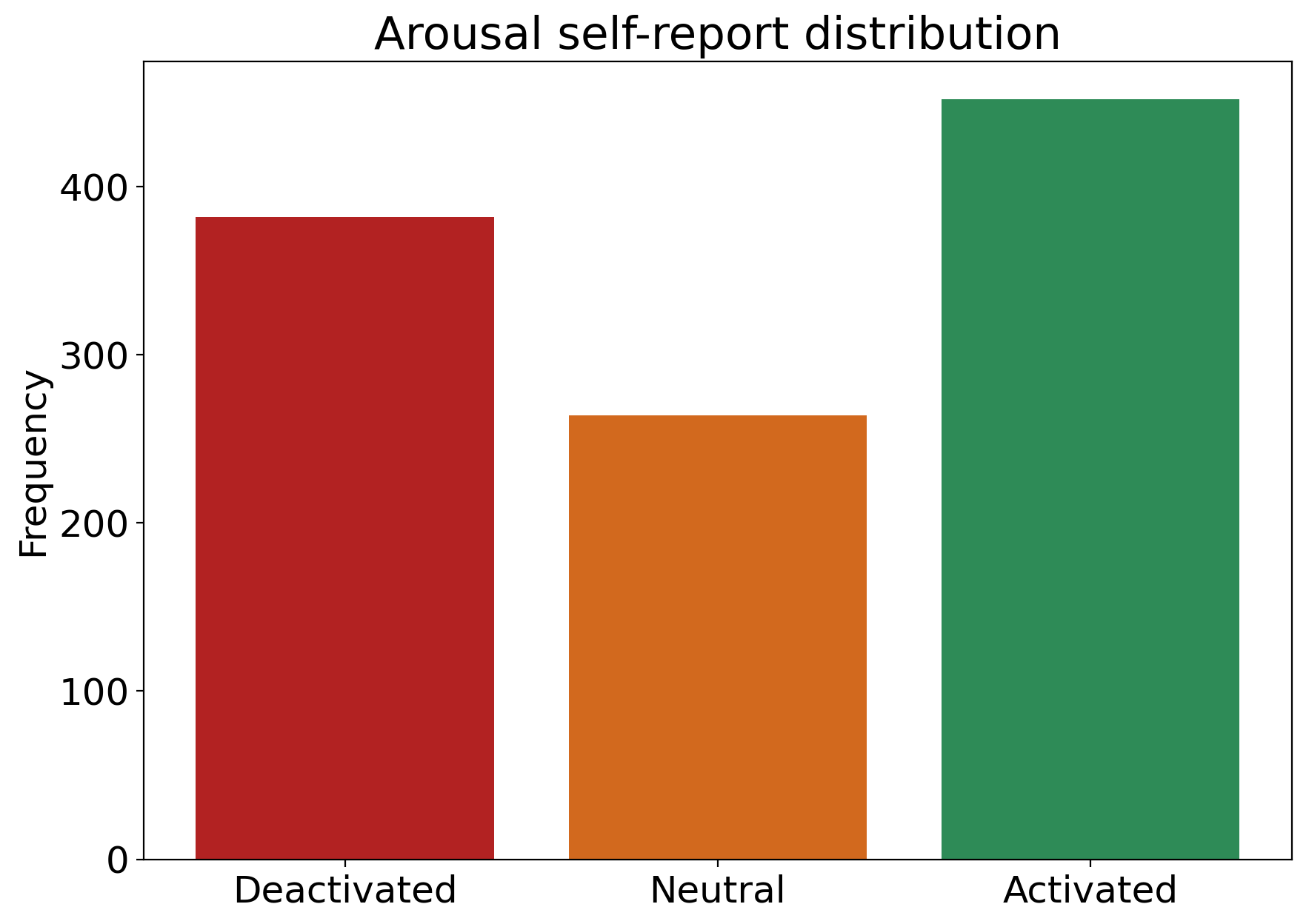}}
  \end{minipage}
  \caption{Categorical self-report annotation distributions. On the left: distributions across the nine quadrants. On the right are three categorical distributions for valence (top) and arousal (bottom).}
  \label{fig:categorical}
\end{figure}

\begin{figure}
    \centering
    \subfloat[Valence mean $\pm$ SD per intended quadrant.]{ \centering
        \includegraphics[width=0.9\linewidth]{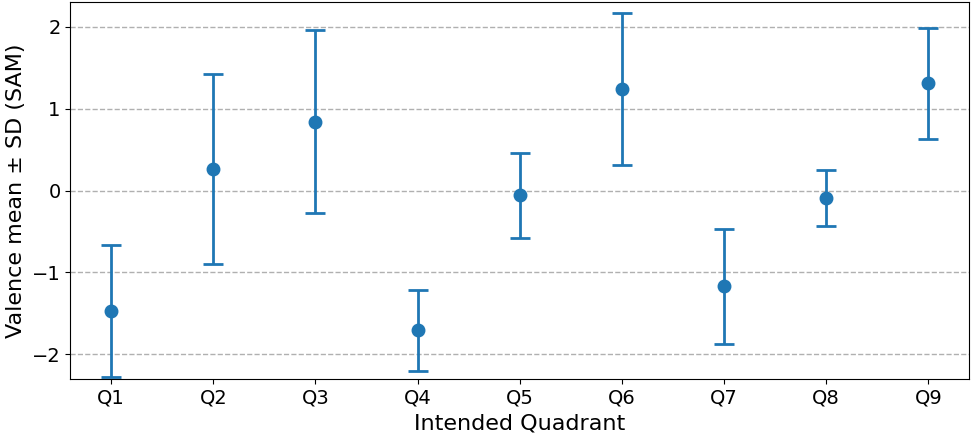}
        \label{fig:valence_sd}
    }\vfill
     \centering
    \subfloat[Arousal mean $\pm$ SD per intended quadrant.]{ \centering
       \includegraphics[width=0.9\linewidth]{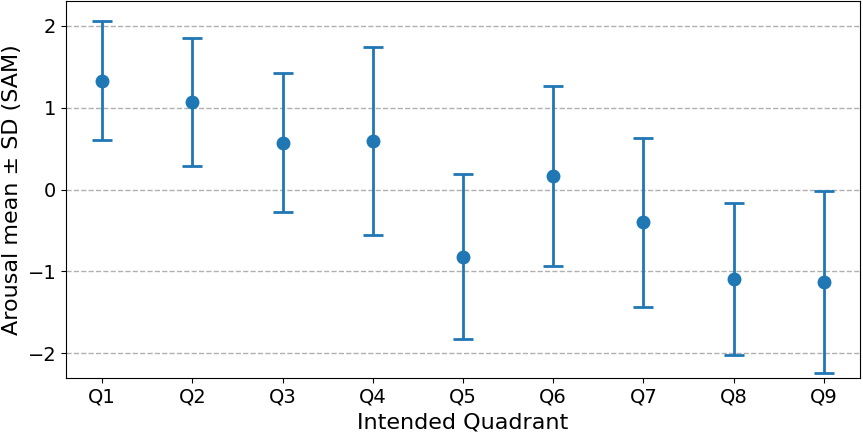}
        \label{fig:arousal_sd}
    } 
    \caption{Mean and standard deviation of (a) valence and (b) arousal ratings across the nine intended quadrants.}
    \label{fig:meanstd_gt}
\end{figure}

\subsection{Self-report annotations}
\label{selfreport}
\begin{figure*}[t]
\centering
\includegraphics[width=170mm,height=80mm]{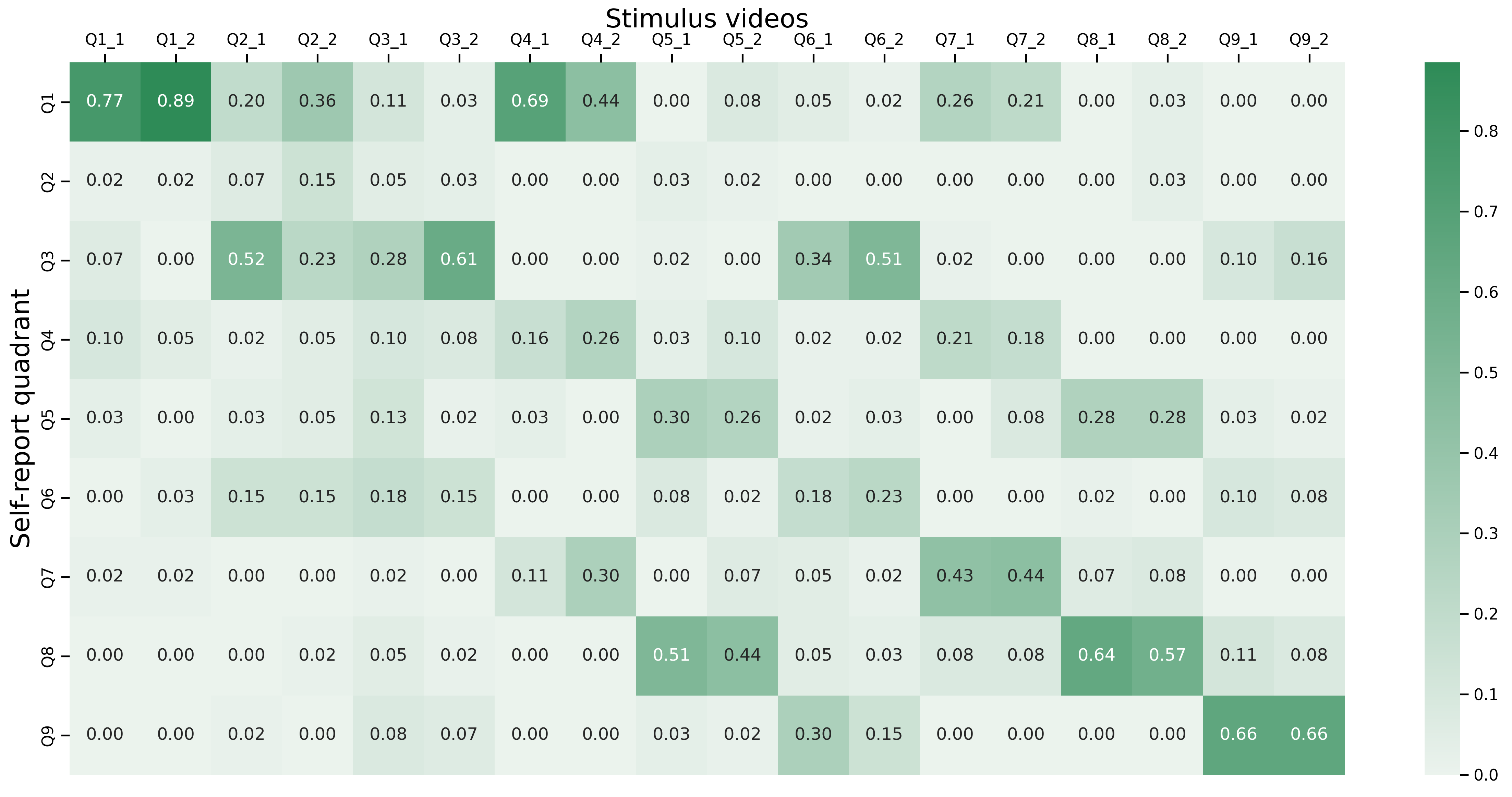}
    \caption{Accuracy heatmap showing the annotations agreement between the elicitated video (x-axis) and the self-report annotations from participants (y-axis).}
    \label{fig:agreement}
\end{figure*}

Figure \ref{fig:annotations} presents the mean self-report results from the CAST-Phys dataset. For each of the nine elicited quadrants, two dots represent the two videos used to stimulate each quadrant. Analyzing the results from 61 participants, we find that the elicitation strategy was effective, leading to a diverse emotional representation. Overall, there is strong agreement across most quadrants; however, variations are observed, particularly concerning arousal. While valence closely aligns with the desired elicitation, some discrepancies in arousal are noted. For example, quadrant seven, representing low arousal and low valence, shows a higher arousal rating similar to quadrant four. Conversely, quadrant five yielded slightly lower arousal ratings from both videos. For the valence dimension, the average responses are generally accurate, except for one video from quadrant three, which produced a more neutral response. Nonetheless, averaging the self-report data can hide key behaviors essential for understanding the emotional response in our dataset. Consequently, we conduct a more in-depth analysis, as shown in Figures \ref{fig:categorical}, \ref{fig:meanstd_gt}  and \ref{fig:agreement}.

Figure \ref{fig:categorical} displays the self-assessment frequencies (number of times that the quadrant was annotated) for the nine elicited quadrants, as well as separate frequencies for valence and arousal. The nine-quadrant distribution reveals an imbalance in the annotations: specifically, quadrant 2 (representing high arousal with neutral valence) is the least populated one, while quadrants 1 and 3 account for the majority of annotations. This suggests that although quadrant 2 successfully elicited the intended arousal response, the valence outcome was split between positive and negative, yielding an average neutral effect as seen in Figure \ref{fig:annotations}. Additionally, quadrant 8 shows the third highest frequency, consistent with the observation that the two videos originally intended for quadrant five were instead annotated as quadrant 8. The rest of the observed quadrants show a very similar distribution. Following \cite{soleymani2011multimodal}, we computed the Cohen's Kappa coefficient for the nine categorical annotations, obtaining an average of 0.3463, similar to \cite{soleymani2011multimodal}. We further analyzed the contributions of valence and arousal separately by categorizing each into three classes: positive, neutral, and negative for valence, and activated, neutral, and deactivated for arousal. In both cases, the neutral state is the least represented one, even though the overall distribution across classes is more balanced than using nine classes, with negative and activated responses being the most common.

To statistically validate the effectiveness of the emotion elicitation protocol, a one-way ANOVA was conducted on participants’ valence and arousal self-reports across the nine intended quadrants. The analysis revealed significant differences for both valence ($F(8, 1089)=250.6,  p < 0.001$) and arousal ($F(8, 1089)= 110.4,  p < 0.001$) across the nine intended quadrants, confirming that the selected stimuli successfully elicited distinct affective responses. Figures \ref{fig:valence_sd} and \ref{fig:arousal_sd} present the mean ± SD valence and arousal ratings per quadrant, highlighting the clear differentiation of emotional responses across the affective circumplex.

Figure \ref{fig:agreement} illustrates the agreement, measured as accuracy, between the eighteen elicitation videos (x-axis) and the self-reported results, which are organized into nine quadrants based on arousal and valence ratings from the participants. The heatmap shows that participants achieved over 60\% agreement for quadrants 1, 8, and 9, and one video in quadrant 3. Although quadrants 4, 5, 6, or 7 do not exhibit clear agreement within a single quadrant, the valence ratings are consistent across participants, with differences mainly in the intensity of the emotion. For example, the two videos in quadrant 6 (designed to elicit high valence with neutral arousal) are predominantly annotated as quadrants 3, 6, and 9, reflecting positive emotions at varying arousal levels. Additionally, one video in quadrant three did not perform as expected, while quadrant 2 shows an inverse trend compared to quadrants 4–7, with most ratings concentrated in quadrants 1, 2, and 3, i.e. areas representing high arousal but differing in valence.  

These results suggest that some emotions are inherently harder to elicit, as individuals may react differently to the same stimuli, even assigning opposite valence or arousal ratings. Factors such as education and cultural background may contribute to the variability in annotation agreement. Moreover, the intensity of emotional responses can differ among users. Nowadays, our society is exposed to a large amount of multimedia content, and many individuals are accustomed to high levels of stimulation, meaning that some may require stronger stimuli to experience emotions more intensely.

\subsection{Multi-modal emotion recognition}
\label{multi_contact}

In this subsection, we present our methodology and analysis for emotion recognition using the recorded modalities. We follow the circumplex model of emotion, adopting a three-class classification for valence (positive, neutral, and negative) and arousal (activated, neutral, and deactivated). To estimate each emotional class, we apply the Analysis of Variance (ANOVA) test for feature selection and utilize machine learning classifiers for class prediction.

To identify the most relevant features for valence and arousal detection, we employ the ANOVA test, a statistical method that compares the variation between group means to the variation within each group. In our work, the ANOVA test is applied to each of the three classes for both valence and arousal, analyzing each feature individually. A significance threshold of p-value 0.05 is used, meaning that any feature with a p-value greater than 0.05 is rejected.

Before applying the ANOVA test, we refine the temporal window to focus on the most relevant period of the stimulus in each video, ensuring meaningful feature extraction. As shown in Figure \ref{fig:modalities}, the intended stimulus does not elicit an amusement response until the 38th second. This is evident in both facial expressions and physiological signals, with increased PPG noise, a longer interbeat interval, rising EDA levels, and a more irregular respiration pattern. During the first 38 seconds, which introduce the stimulus, the participant remains neutral. To mitigate averaging effects, shorter windows are used, particularly for videos with high-arousal stimuli.

After applying the ANOVA test, we identify 26 meaningful features for valence and 15 for arousal:

\begin{itemize}
    \item For the valence dimension, the PPG modality yields two temporal features (heart rate standard deviation and SDNN), two frequency features (LF and the LF/HF ratio), and three non-linear features (TINN, SD2, and the Poincaré ratio between SD1 and SD2). The respiration signal produces significant results for all features except the mean of respiration derivation. In the EDA modality, only the SCL mean and SCR mean of peak magnitudes are not significant. Facial expressions provide 11 meaningful features, including AU02, AU04, AU06, AU10, AU12, AU14, AU15, AU17, AU25, AU26, and AU28.

    \item For the arousal dimension, the PPG modality retains only two significant features, LF and the LF/HF ratio. The EDA modality contributes four significant features: SCL and SCR standard deviations, SCR mean, and the mean of peak magnitudes. Facial expressions are the most significant modality, including AU01, AU06, AU09, AU12, AU14, AU17, AU24, AU25, and AU26. The respiration signal does not yield any significant features for valence. However, the respiration mean is considered only for single-modality computation, as its p-value of 0.067 is close to 0.05, approaching the margin of significance.
    
\end{itemize}

\begin{figure}[b ]
\centering
\includegraphics[width=89mm,height=50mm]{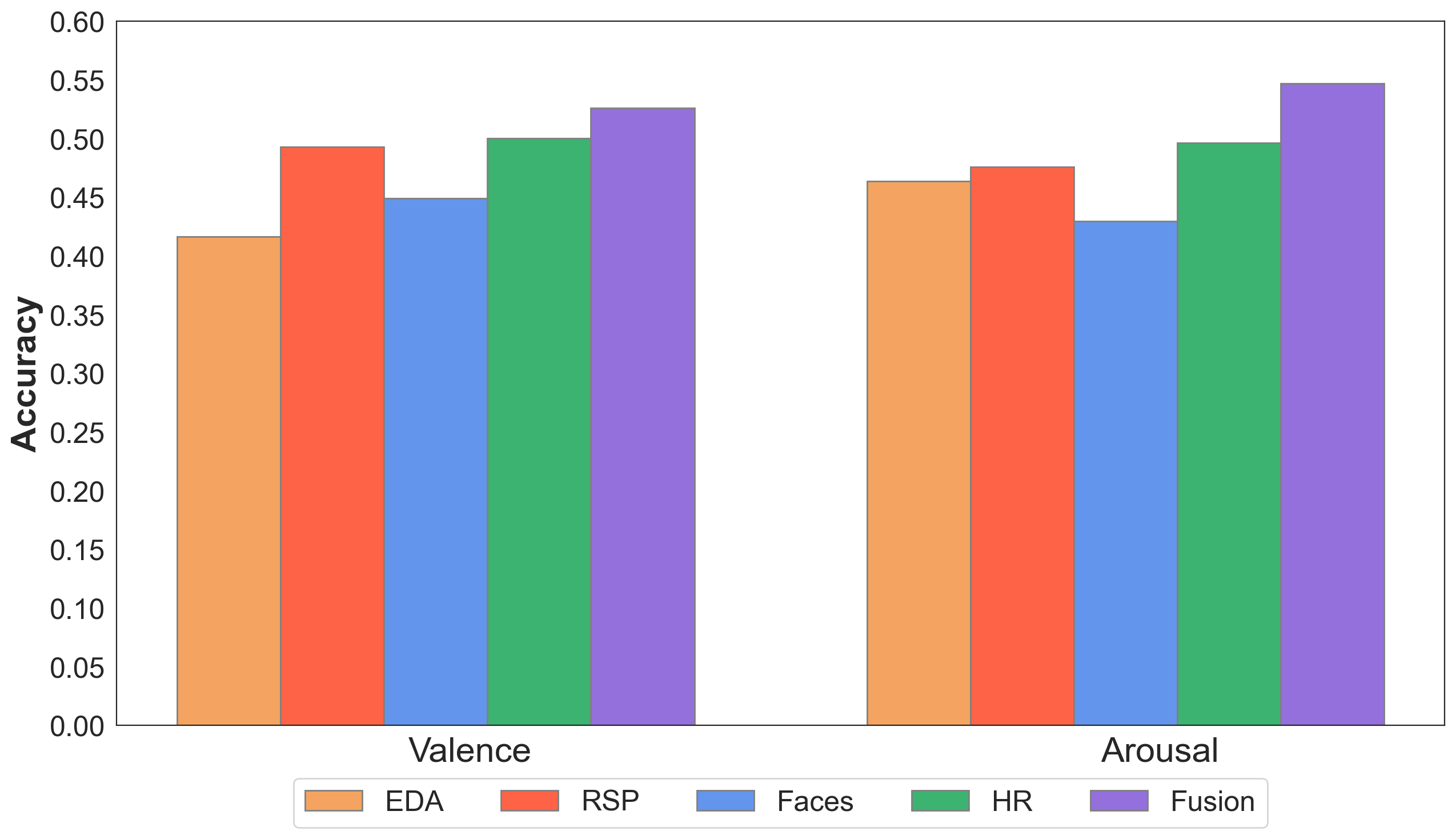}
    \caption{Valence and Arousal accuracy classification results for each modality.}
    \label{fig:modality_contact}
\end{figure}

For emotion classification, we evaluate eight machine learning classifiers: K-Nearest Neighbors (KNN), Support Vector Machine (SVM) with linear and Radial Basis Function (RBF) kernels, Decision Tree, Random Forest, AdaBoost, Naive Bayes, and Quadratic Discriminant Analysis (QDA). We assess their performance using a six-fold cross-validation strategy, measuring accuracy and F1-score across the three classes for both valence and arousal.

Figure \ref{fig:modality_contact} presents the emotion recognition results, showing the best accuracy for each modality, as well as their fusion for both arousal (on the left) and valence (on the right). Regarding single-modality performance, for valence dimension, heart rate and respiration modalities demonstrate the highest performance, with 50.1\% accuracy using AdaBoost and 49.3\% accuracy using Random Forest, respectively. Facial expression features achieve 45\% accuracy with SVM, while EDA features reach 41\% accuracy with Random Forest. For the arousal dimension, the heart rate features reach 49.6\% accuracy with Adaboost, followed by respiration and EDA, with 47.7 \% and 47.2\%, respectively. Finally, faces obtain a performance of 42.9 \% with Naive Bayes.

Surprisingly, facial expressions yielded a low valence classification performance. This result is unexpected, given that facial expressions are strong indicators of valence and previously exhibited the highest number of meaningful features. To investigate potential errors in the AUs model, we conducted a manual annotation experiment where three external annotators independently labelled the same frames used for the facial model. The annotators achieved a valence accuracy of 45\%, closely matching the model's performance.

\begin{figure}[t]
    \centering
    \subfloat{%
        \includegraphics[width=0.10\textwidth, height=0.75in]{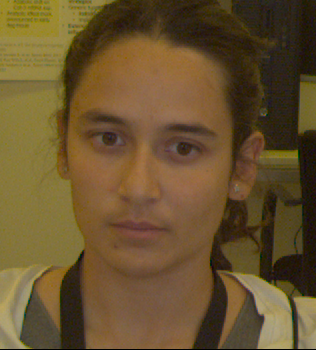}
        \label{fig:sub15}
    }\hfill
    \subfloat{%
        \includegraphics[width=0.10\textwidth, height=0.75in]{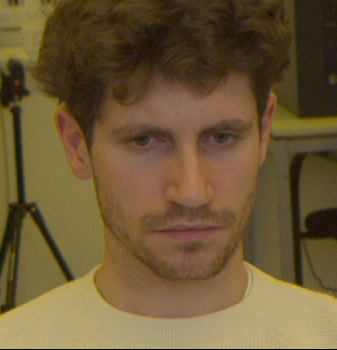}
        \label{fig:sub50}
    }\hfill
    \subfloat{%
        \includegraphics[width=0.10\textwidth, height=0.75in]{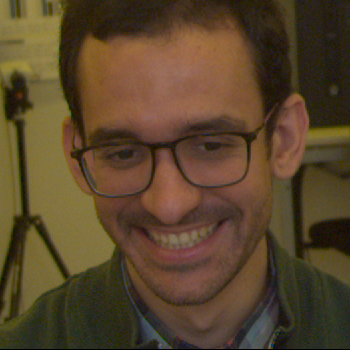}
        \label{fig:sub55}
    }\hfill
    \subfloat{%
        \includegraphics[width=0.10\textwidth, height=0.75in]{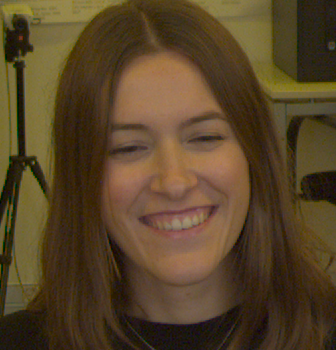}
        \label{fig:sub32}
    }
    \caption{Facial expressions of four subjects during stimulus $Q3\_2$ at the same moment. All subjects reported positive and activated emotional responses ($Q3$).}
    \label{fig:four_faces}
\end{figure}

These findings suggest that, in realistic scenarios, facial expressions may not always provide the most relevant emotional information, whereas subtle physiological cues might be more informative. This is further illustrated in Figure \ref{fig:four_faces}, which displays four participants during the elicitation with the second video of quadrant 3. As shown previously in Figure \ref{fig:agreement}, this stimulus achieved a self-report agreement of 61\%. Although all four participants labelled the video as belonging to quadrant Q3, their facial expressions varied significantly. The first two participants exhibited neutral expressions, whereas the last two appeared amused, highlighting the variability in facial cues despite shared emotional labelling. Furthermore, these results support the conclusions from Subsection \ref{selfreport}, suggesting that extensive exposure to multimedia content can reduce the dominance of emotional expressions when using video clips for elicitation.

Finally, the fusion of modalities (represented in magenta) significantly enhances performance compared to single-modality approaches for both valence and arousal. By integrating all four recorded modalities, we achieve an accuracy of 52.6\% for valence and 54.7\% for arousal, representing an improvement of 2.5\% and 5\%, respectively, over the best-performing single modality. For a more detailed analysis, Table \ref{table:contact_6fold_VA_ML} presents the six-fold cross-validation results for emotion classification across different classifiers within the fusion-based model. The findings indicate that Random Forest achieves the highest accuracy and F1-score for both valence and arousal, followed closely by AdaBoost and Decision Tree classifiers.

\begin{table}[b]
  \caption{Multi-modal emotion recognition 6-Fold Average Performance}
  \adjustbox{width=0.49\textwidth}{
  \renewcommand{\arraystretch}{1.35}
  \centering
  \begin{tabular}{c|cc|cc}
  \hline
     \multirow{2}{2cm}{\centering Method} &\multicolumn{2}{c}{Valence} &\multicolumn{2}{|c}{Arousal}\\ 
    \cline{2-5}
    & Accuracy$\uparrow$ & F1-score$\uparrow$& Accuracy$\uparrow$& F1-score$\uparrow$\\
    \hline
    Nearest Neighbors & 0.434 & 0.421 & 0.415 & 0.398  \\  
    Linear SVM & 0.474 &  0.419 & 0.488 &  0.417 \\
    RBF SVM & 0.396 & 0.225 & 0.409 & 0.252 \\
    Decision Tree & 0.484 & 0.461 & 0.525 & 0.486 \\
    Random Forest & \textbf{0.526} & \textbf{0.497} &\textbf{0.547} & \textbf{0.476} \\
    AdaBoost & 0.510 & 0.495 & 0.521 & 0.475 \\
    Naive Bayes & 0.429 & 0.431 & 0.466 & 0.42 \\
    QDA & 0.462 & 0.439 & 0.461 & 0.423 \\
\hline
  \end{tabular}
  }
  \label{table:contact_6fold_VA_ML}
\end{table}

\subsection{Remote physiological signal extraction}
\label{remote_ext}

Considering the results and importance of heart rate and respiration modalities in the emotion recognition of the previous subsection, we decide to extract these two signals remotely due to their potential link with emotional cues. To extract physiological signals from facial videos, we compare three state-of-the-art methods: two hand-crafted approaches, ICA \cite{poh2010advancements} and POS \cite{wang2016algorithmic}, and one data-driven method, TDM \cite{comas2022efficient}. The TDM model is a lightweight spatiotemporal network that effectively estimates rPPG spatiotemporal features by aggregating temporal derivative modules (TDM), emulating a Taylor series expansion. All these methods are chosen for their versatility, efficiency and suitability for real-time applications while maintaining competitive heart rate (HR) estimation performance. Table \ref{table:rPPG} presents the six-fold average performance of each method using common remote HR and respiration rate (RR) estimation metrics, including mean absolute error (MAE), root mean squared error (RMSE), and Pearson correlation ($\rho$). The results indicate that the hand-crafted models yield acceptable performance, with the POS method achieving an error of 2.01 BPM across the entire dataset. However, the data-driven TDM approach outperforms both, obtaining the lowest HR error with an MAE of 1.1 BPM, an RMSE of 4.51 BPM, and a correlation of 0.93. Additionally, although the original TDM model was designed exclusively for rPPG signal recovery, we enhanced it by, incorporating two final fully connected layers, yielding a multi-task regression framework that simultaneously estimates both rPPG and rRSP signals. The model was trained on the CAST-Phys dataset following the same six-fold cross-validation strategy described in Subsection \ref{multi_contact}. For remote respiration estimation, we achieved baseline results of 2.21 RPM MAE and 3.25 RPM RMSE. 

Table \ref{table:TDM_fold} provides detailed results across the six folds, showing that the TDM approach achieves consistently high accuracy in HR estimation, with MAE ranging from 0.72 to 1.71 BPM and strong correlations ($\rho > 0.87$), particularly in Fold 4 ($\rho = 0.98$). RR estimation exhibits greater variability, with MAE between 1.89 and 2.71 RPM and moderate correlations ($\rho = 0.19–0.53$). Folds 1 and 4 yield the most stable performance for both HR and RR, while Folds 2, 3, and 5 show slightly degraded HR and RR correlations. Overall, these results demonstrate the robustness of the TDM method for HR extraction and its moderate sensitivity in RR estimation to inter-subject differences.

\begin{figure}[b]
\centering
\includegraphics[width=89mm,height=55mm]{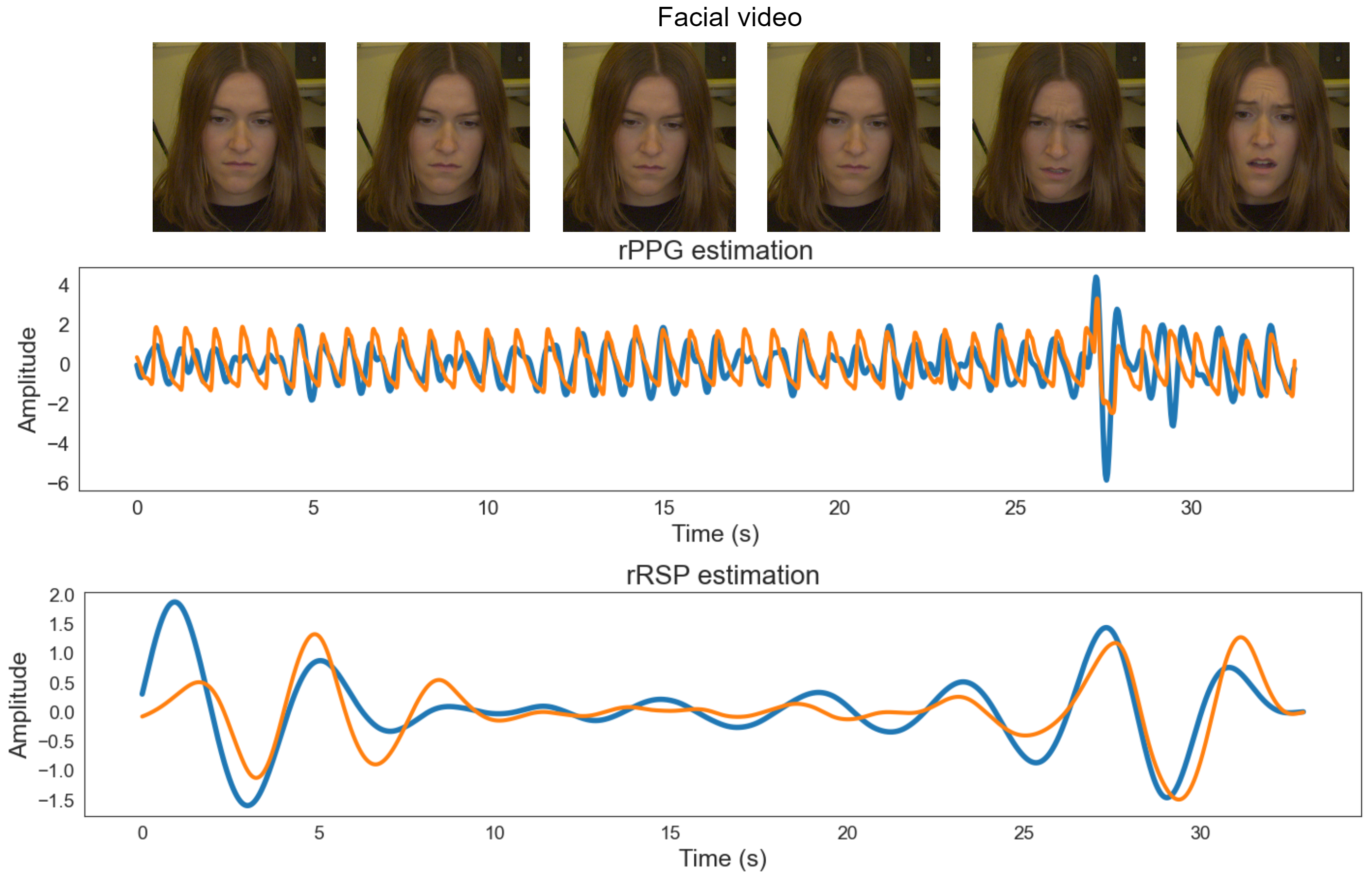}
    \caption{Remote physiological extraction of a participant during quadrant Q1 elicitation. From top to bottom: key facial video frames, estimated rPPG, and respiration signals. Predictions are shown in blue, while contact-based signals are in orange.}
    \label{fig:remote_ext}
\end{figure}

Given its superior performance and ability to extract both PPG and respiration signals, we employ the TDM model to obtain the remote physiological signals used for emotion recognition, as detailed in the next subsection. Figure \ref{fig:remote_ext} depicts an example of our remote physiological extraction during the high-arousal elicitation observed at the end of the video.

\begin{table}[t]
  \caption{Remote Pulse and respiration measurement on the CAST-Phys.}
  \label{table:rPPG}
  \begin{threeparttable}
  \renewcommand{\arraystretch}{2}
  \centering
  \adjustbox{width=0.49\textwidth}{
  \begin{tabular}{c|ccc|ccc}
    \hline
     \multirow{2}{2cm}{\centering Method} &\multicolumn{3}{c}{Heart rate (in BPMs)} &\multicolumn{3}{|c}{Respiration rate (in RPMs)}\\ 
    \cline{2-7}
    &MAE$\downarrow$&RMSE$\downarrow$&$\rho$$\uparrow$&MAE$\downarrow$&RMSE$\downarrow$&$\rho$$\uparrow$\\  
    \hline
    ICA \cite{poh2010advancements}  & 6.15 & 11.62 & 0.52 & - & - & - \\ 
    POS \cite{wang2016algorithmic}  & 2.01 & 5.47 & 0.89 & - &  - &  -\\   
    TDM \cite{comas2022efficient}  & 1.11 & 4.51 & 0.93 & 2.21 & 3.25 & 0.33 \\
    \hline
  \end{tabular}
  }
    \begin{tablenotes}
    \item \centering \footnotesize MAE: Mean Absolute Error, RMSE: Root Mean Squared Error,\\ $\rho$: Pearson Correlation
    \end{tablenotes}
  \end{threeparttable}
\end{table}

\begin{table*}[t]
  \caption{Remote heart rate and respiration estimation on the CAST-Phys using TDM approach for each fold.}
  \label{table:TDM_fold}
  \begin{threeparttable}
  \renewcommand{\arraystretch}{1.5}
  \centering
  \adjustbox{width=0.8\textwidth}{
  \begin{tabular}{c|ccc|ccc}
    \hline
     \multirow{2}{2cm}{\centering Fold} &\multicolumn{3}{c}{Heart rate (in BPMs)} &\multicolumn{3}{|c}{Respiration rate (in RPMs)}\\ 
    \cline{2-7}
    &MAE$\downarrow$&RMSE$\downarrow$&$\rho$$\uparrow$&MAE$\downarrow$&RMSE$\downarrow$&$\rho$$\uparrow$\\  
    \hline
    Fold 1  & 0.72 $\pm$ 0.27 & 3.82 $\pm$ 11.83 & 0.94 $\pm$ 0.02 & 1.89 $\pm$ 0.15 & 2.82 $\pm$ 0.98 & 0.53 $\pm$ 0.06\\
    Fold 2  & 1.01 $\pm$ 0.31 & 4.32 $\pm$ 13.38 & 0.87 $\pm$ 0.04 & 2.31 $\pm$ 0.20 & 3.54 $\pm$ 1.72 & 0.23 $\pm$ 0.07\\
    Fold 3  & 0.95 $\pm$ 0.29 & 4.07 $\pm$ 12.72 & 0.94 $\pm$ 0.03 & 2.71 $\pm$ 0.18 & 3.63 $\pm$ 1.38 & 0.23 $\pm$ 0.07 \\
    Fold 4  & 0.73 $\pm$ 0.13 & 1.88 $\pm$ 1.07 & 0.98 $\pm$ 0.01 & 1.93 $\pm$ 0.17 & 3.04 $\pm$ 1.23 & 0.49 $\pm$ 0.07\\
    Fold 5  & 1.71 $\pm$ 0.52 & 7.16 $\pm$ 25.08 & 0.92 $\pm$ 0.03 & 2.30 $\pm$ 0.18 & 3.38 $\pm$ 1.47 & 0.19 $\pm$ 0.07 \\
    Fold 6  & 1.51 $\pm$ 0.42 & 5.80 $\pm$ 17.83 & 0.94 $\pm$ 0.03 & 2.11 $\pm$ 0.17 & 3.08 $\pm$ 1.13 & 0.33 $\pm$ 0.07 \\
    \hline
    Average  & 1.11 $\pm$ 0.32 & 4.51 $\pm$ 13.65 & 0.93 $\pm$ 0.03 & 2.21 $\pm$ 0.18 & 3.25 $\pm$ 1.32 & 0.33 $\pm$ 0.07\\
    \hline
  \end{tabular}
  }
    \centering
    \begin{tablenotes}
    \item \centering \footnotesize MAE: Mean Absolute Error, RMSE: Root Mean Squared Error, $\rho$: Pearson Correlation
    \end{tablenotes}
  \end{threeparttable}
\end{table*}

\subsection{Remote multi-modal emotion recognition}
\label{multi_remo}

Following Subsection \ref{multi_contact}, this subsection focuses on emotion recognition using statistical ANOVA tests and machine learning. However, instead of contact-based modalities, we rely entirely on remote modalities, including facial expressions, rPPG, and rRSP signals. For facial expressions, we use the same features as in Subsections \ref{preprocessing} and \ref{multi_contact}. In contrast, for physiological signals, we extract the features described in \ref{preprocessing} but computed from rPPG and rRSP signals derived from the facial videos.

The selected features closely resemble those identified in Subsection \ref{multi_contact}. For the valence dimension, heart rate yields four significant features: pNN50, heart rate standard deviation, LF, and the LF/HF ratio, while respiration retains the same three meaningful features as before. Regarding arousal, heart rate includes four relevant features: LF, LF/HF ratio, TINN, and the Poincaré ratio between SD1 and SD2. Additionally, respiration now contributes two meaningful features: respiration rate mean and standard deviation.

Notably, heart rate frequency-domain features appear to be robust indicators of emotional states, regardless of whether they are derived from contact-based or remote estimation methods. Meanwhile, the respiration rate mean emerges as the most stable feature for the respiration signal. The variation in significant features obtained from contact-based signals may be attributed to subtle differences in the extracted physiological data.

As shown in Table \ref{table:rPPG}, the six-fold average results of remote HR indicate good performance when evaluating HR in the whole video protocol. However, differences between contact-based and remote signals remain. Heart rate variability computation is particularly sensitive to erroneous peak detections, which can occur in remote photoplethysmography (rPPG) signals due to motion artifacts. Some participants viewed the videos under realistic conditions, leading to challenges such as facial occlusions (e.g. covering the face in response to disgust), rapid head movements, or difficult head positions. These factors may have compromised rPPG estimation, leading to some inaccurate peak detections and affecting heart rate variability analysis. In contrast, respiration rate estimation, which relies solely on the mean, is less susceptible to such errors.

\begin{figure}[t]
\centering
\includegraphics[width=89mm,height=50mm]{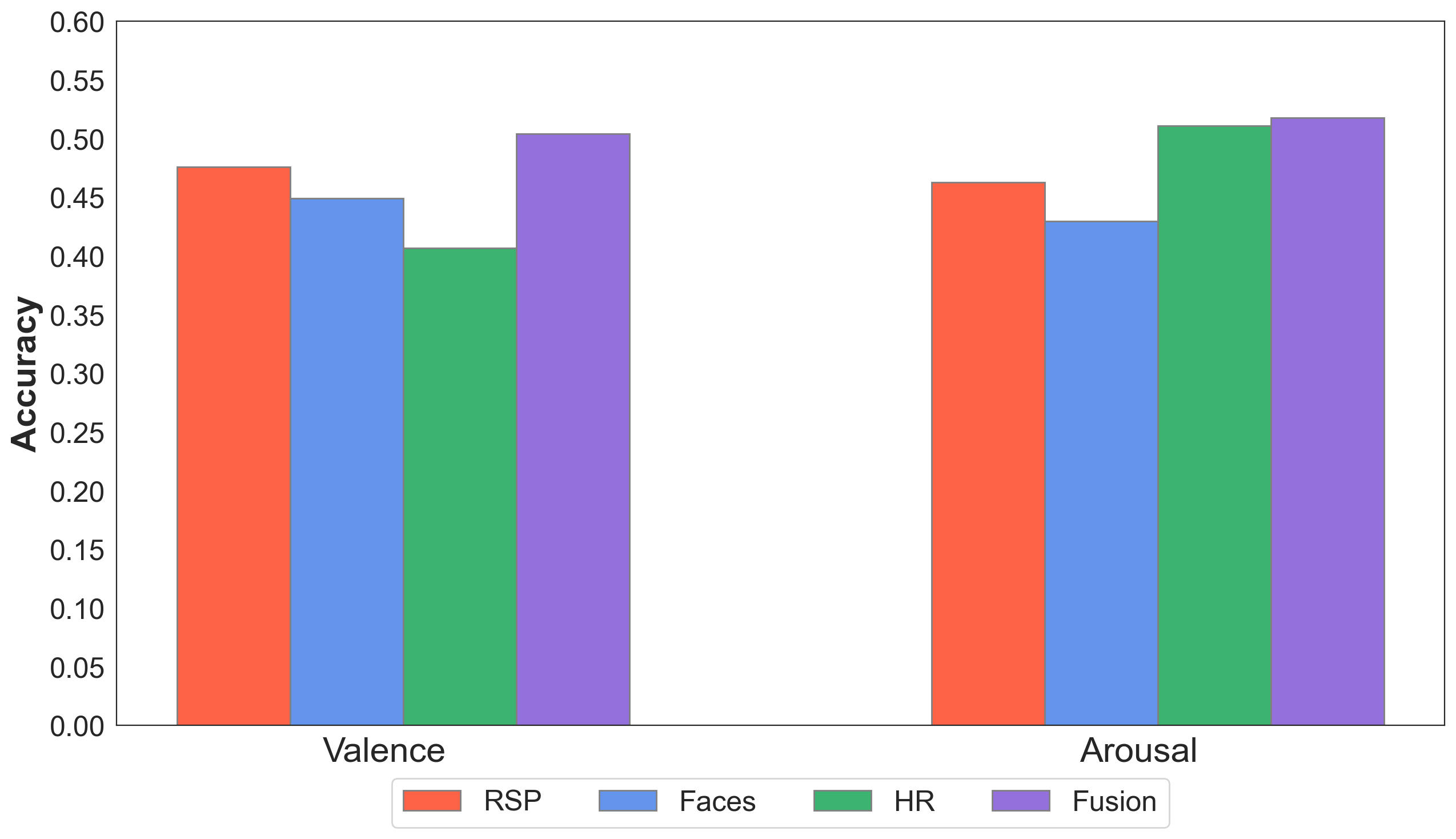}
    \caption{Valence and Arousal accuracy classification results for each remote extracted modality.}
    \label{fig:modality_remote}
\end{figure}

Figure \ref{fig:modality_remote} presents the accuracy performance of remote emotion recognition for each modality. For valence, heart rate exhibits a notable performance drop, achieving only 40.7\% accuracy with Random Forest compared to the contact-based signal. In contrast, respiration maintains a similar performance, reaching 47.6\% accuracy with Random Forest, closely matching the contact-based results. For arousal, the performance remains nearly unchanged between remote and contact-based modalities. Heart rate achieves 51.1\% accuracy using Random Forest, slightly outperforming the 49.6\% recorded in the contact-based setup. Similarly, respiration attains 46.3\% accuracy with Random Forest, compared to 47.6\% in the contact-based condition. These findings suggest that remote emotion recognition can achieve comparable results to contact-based methods, with the primary discrepancy observed in valence estimation using remote heart rate features.

\begin{table}[b]
  \caption{Remote multi-modal emotion recognition 6-Fold Average Performance}
  \adjustbox{width=0.49\textwidth}{
  \renewcommand{\arraystretch}{1.35}
  \centering
  \begin{tabular}{c|cc|cc}
  \hline
     \multirow{2}{2cm}{\centering Method} &\multicolumn{2}{c}{Valence} &\multicolumn{2}{|c}{Arousal}\\ 
    \cline{2-5}
    & Accuracy$\uparrow$ & F1-score$\uparrow$& Accuracy$\uparrow$& F1-score$\uparrow$\\
    \hline
    Nearest Neighbors & 0.404 & 0.395 & 0.370 & 0.351  \\  
    Linear SVM & 0.477 &  0.431 & 0.480 &  0.409 \\
    RBF SVM & 0.396 & 0.225 & 0.409 & 0.244 \\
    Decision Tree & \textbf{0.505} & \textbf{0.496} & 0.455 & 0.427 \\
    Random Forest & 0.504 & 0.476 &\textbf{0.518} & \textbf{0.452} \\
    AdaBoost & 0.485 & 0.470 & 0.49 & 0.443 \\
    Naive Bayes & 0.455 & 0.453 & 0.454 & 0.408 \\
    QDA & 0.426 & 0.424 & 0.445 & 0.413 \\
\hline
  \end{tabular}
  }
  \label{table:remote_6fold}
\end{table}

The drop in valence accuracy for remote heart rate can be attributed to the fewer features extracted by ANOVA, particularly the absence of important non-linear features that contribute to higher accuracy. In contrast, arousal performance benefits from the same two frequency-domain features used in the contact-based approach, along with two additional non-linear features, leading to a 1.5\% improvement in accuracy. Hence, these results also highlight the critical role of feature selection in emotion recognition, as heart rate variability non-linear features appear to significantly impact both valence and arousal estimation. Exploring alternative feature selection methods or employing data-driven feature selection techniques could further enhance these baseline results.

\begin{figure*}[t]
    \centering
    \subfloat[Valence multi-modal contact-based]{ \centering
        \includegraphics[width=0.38\textwidth, height=2.25in]{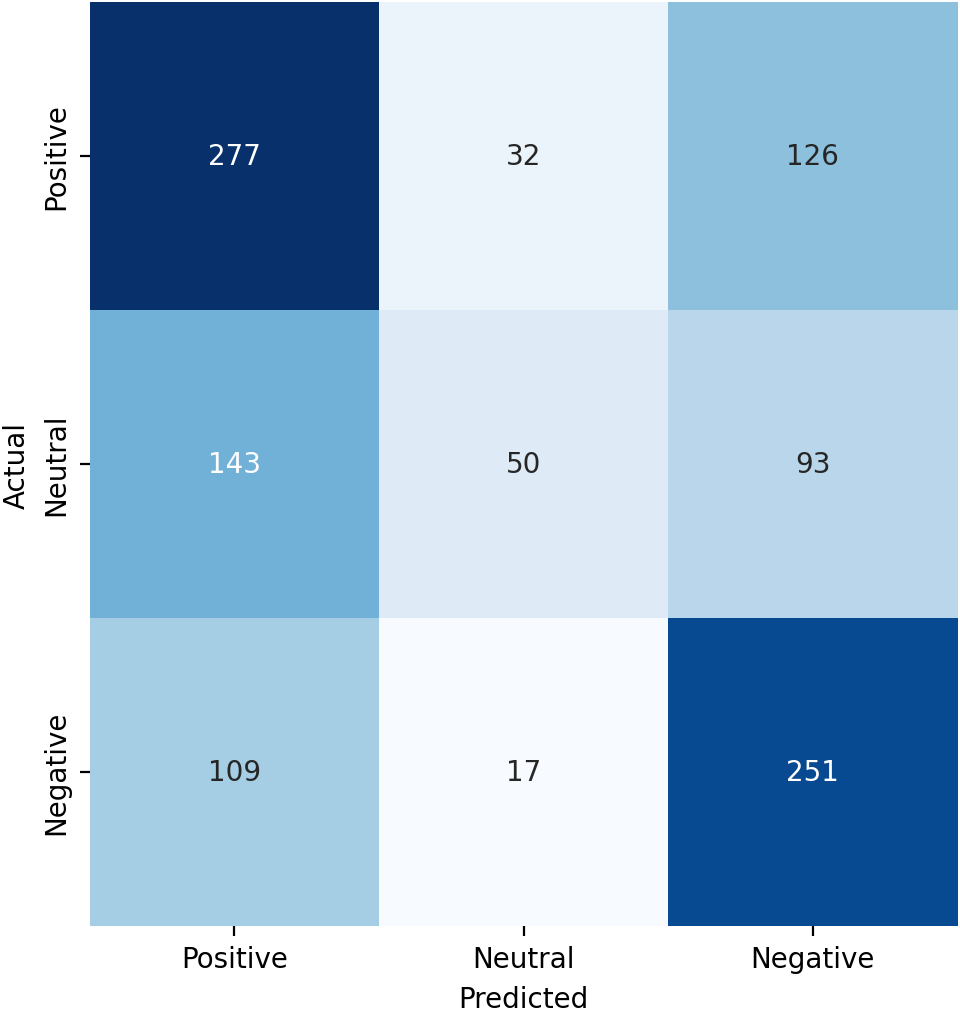}
        \label{fig:vcm_all_contact}
    }\hfill
     \centering
    \subfloat[Valence multi-modal remote-based]{ \centering
       \includegraphics[width=0.38\textwidth, height=2.25in]{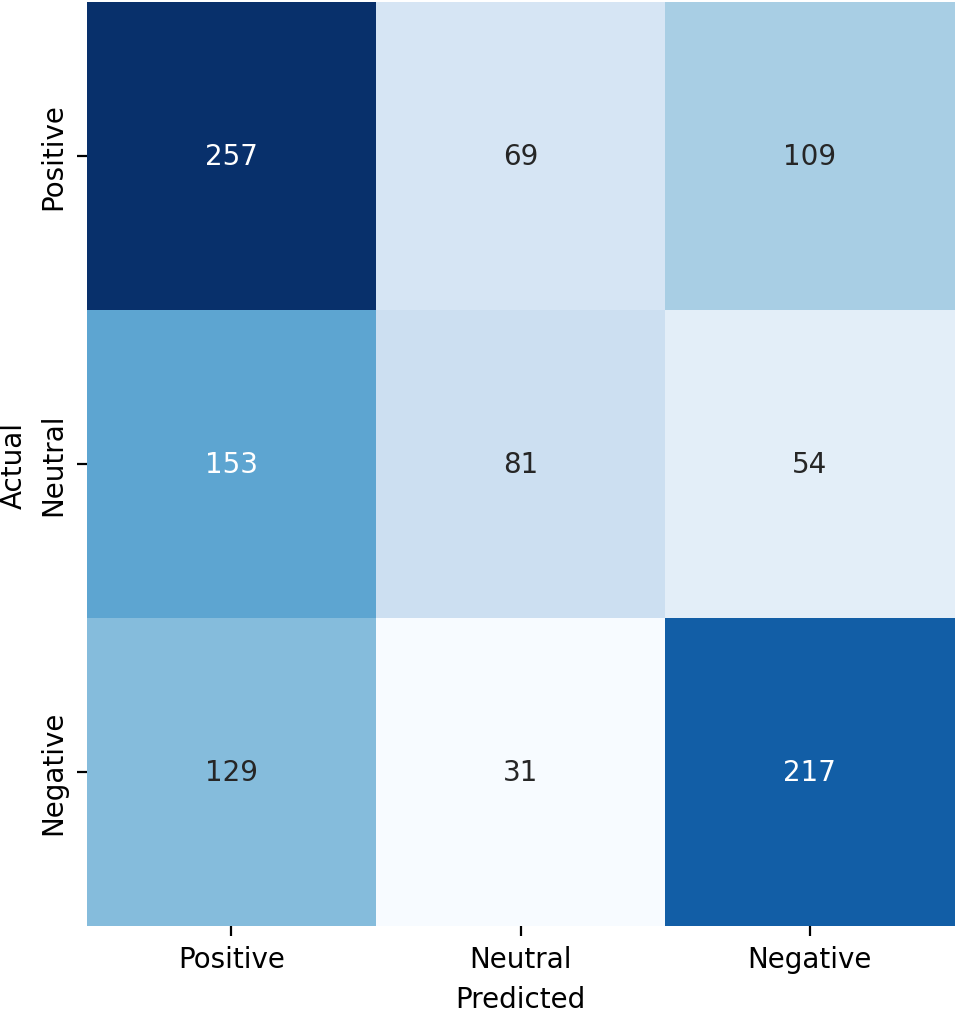}
        \label{fig:vcm_all_remote}
    }
    \vfill
     \centering
    \subfloat[Arousal multi-modal contact-based]{ \centering
        \includegraphics[width=0.38\textwidth, height=2.25in]{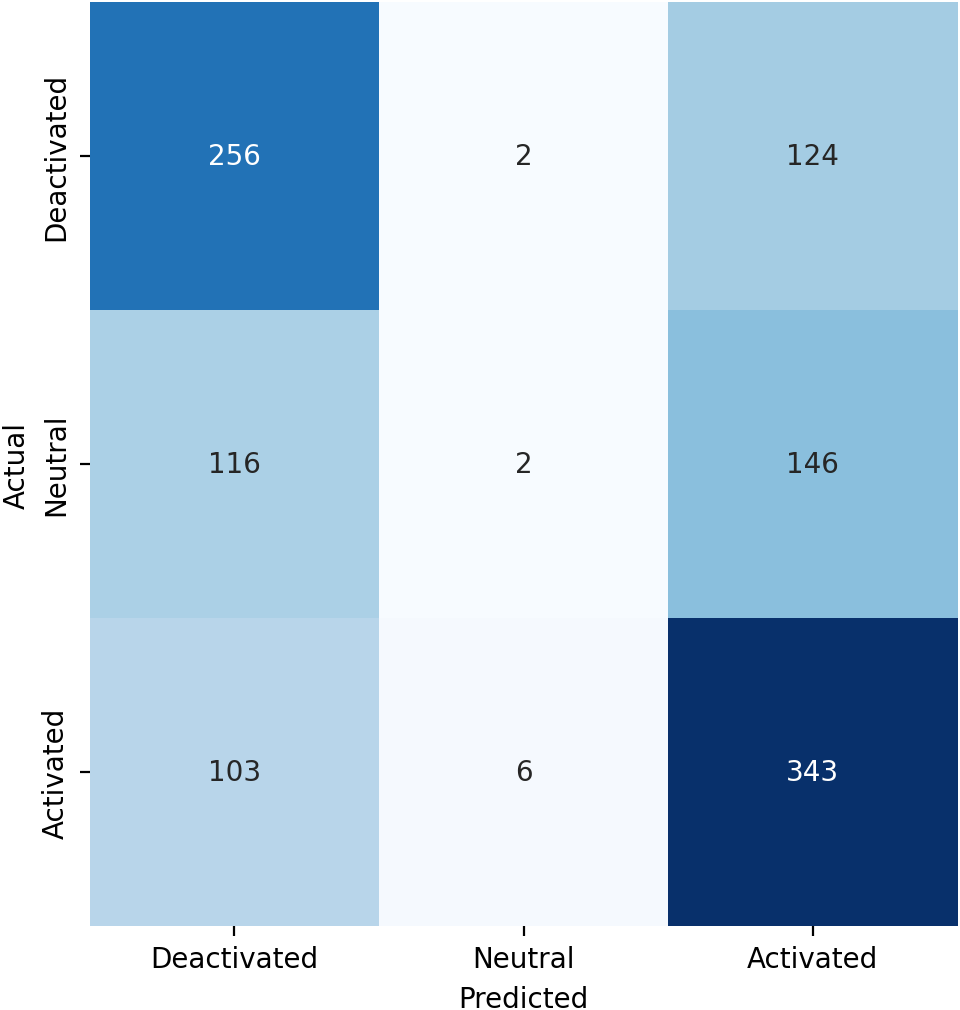}
        \label{fig:acm_all_contact}
    }\hfill
     \centering
    \subfloat[Arousal multi-modal remote-based]{ \centering
        \includegraphics[width=0.38\textwidth, height=2.25in]{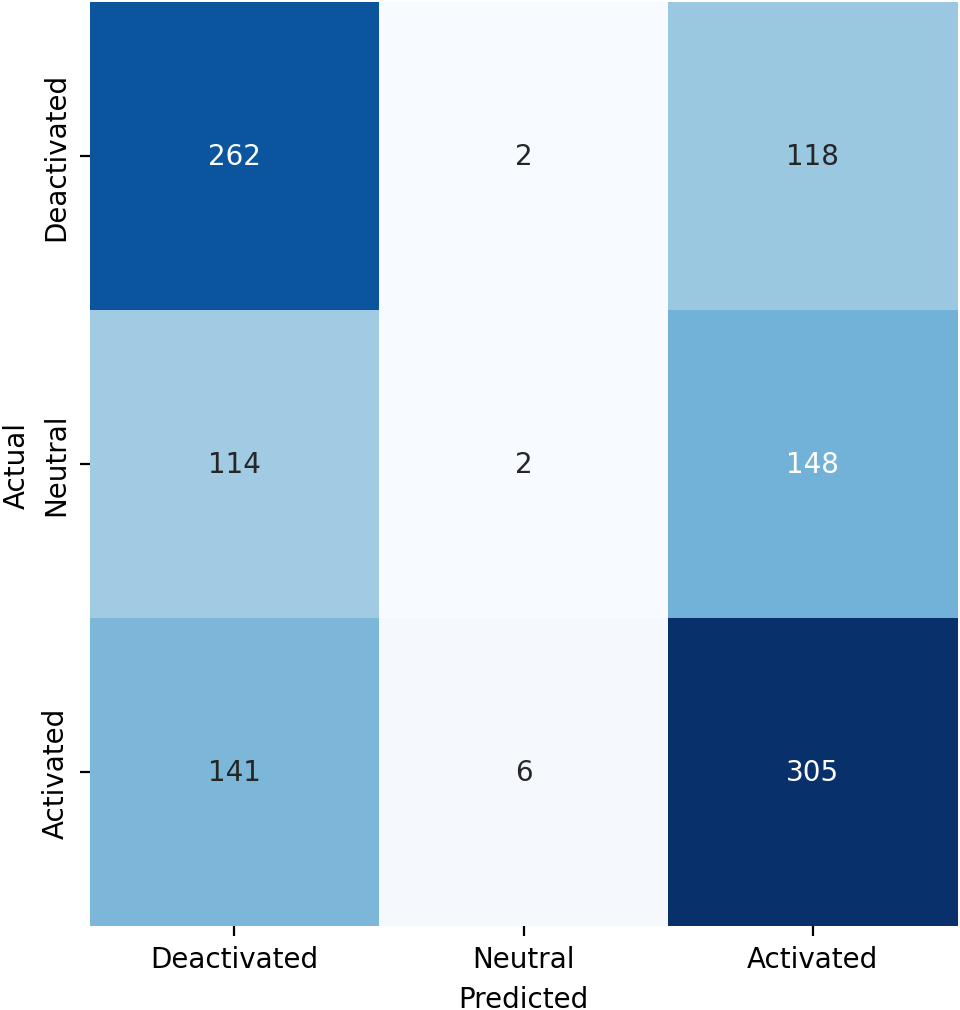}
        \label{fig:acm_all_remote}
    }
    
    \caption{Valence and Arousal confusion matrices comparison between multimodal contact-based and remote-based.}
    \label{fig:cm}
\end{figure*}

As shown in Subsection \ref{multi_contact}, the combination of the extracted modalities, in this case with facial expressions, rPPG and rRSP, reflects in Figure \ref{fig:modality_remote} and Table \ref{table:remote_6fold} a superior performance than each modality individually. From Table \ref{table:remote_6fold}, we appreciate that Random Forest, Decision Tree and Adaboost seem to be the most robust classifiers for the evaluated modalities like in contact-based estimation. Comparing the multi-modal emotion recognition results obtained from contact and fully remote modalities, can observe very similar performance with a slight decrease in remote modality. Figure \ref{fig:cm} presents the confusion matrices for valence and arousal classification across different cases. For arousal (Figures \ref{fig:acm_all_contact} and \ref{fig:acm_all_remote}), we observe a common issue in both contact-based and remote approaches: the neutral state is not well detected, and arousal tends to be polarized between deactivated and activated emotions. This aligns with the findings discussed in previous subsections, where the presence of neutral states within activated and deactivated emotions makes accurate classification of this category more challenging. In contrast, for valence (Figures \ref{fig:vcm_all_contact} and \ref{fig:vcm_all_remote}), the neutral class achieves significantly better classification in both cases, with even stronger performance in remote emotion recognition. Although remote emotion recognition shows slightly lower accuracy for positive and negative states compared to contact-based methods, it appears to achieve a more balanced classification across the three arousal categories.

\section{Discussion and future work}
\label{discussion_future}

During the acquisition and analysis of the collected dataset, several key conclusions emerged. First, emotion elicitation proved to be one of the most challenging and critical aspects of this work. Despite efforts to design realistic stimuli that were long enough to evoke emotions but short enough to prevent boredom, achieving genuine emotional responses through multimedia content alone remained difficult. The vast exposure to social media further complicates the extraction of dominant and authentic emotions. While a more personalized elicitation approach, incorporating psychological support, could provide more consistent and robust emotional responses, this would require significantly more resources and may raise other issues regarding the homogeneity of the obtained reactions.

Our results also indicate that facial expressions do not always provide significant emotional information, as the expressiveness of participants differed greatly even when experiencing the same emotion. In contrast, physiological signals such as heart rate and respiration rate seem to offer more robust insights for emotion recognition, as these signals are more likely to escape voluntary control compared to facial expressions.

Additionally, this work demonstrates the comparable performance of contact-based and fully remote signal extraction for emotion recognition, particularly in terms of multi-modal fusion. This physiological signal extraction opens up many potential solutions in emotion recognition, such as providing implicit support to facial video-based emotion recognition or reducing the intrusion for contact-based devices, thereby improving the comfort of participants during emotion elicitation.

The CAST-Phys database offers numerous opportunities for future research. In this study, we provide baseline results using established hand-crafted features and the ANOVA test for feature selection, which are commonly used in emotion recognition studies. However, exploring data-driven techniques could yield more refined fusion between modalities, automatically identifying the most relevant features or automatically and adaptively selecting key moments from recorded videos.

In terms of remote physiological signal sensing, we adopted an efficient state-of-the-art model. Although achieving the best possible recovery accuracy was not the primary focus of this work, future research could explore novel rPPG approaches capable of estimating physiological signals under some of the challenging and realistic conditions included in this database, such as facial occlusions and rapid head movements. Moreover, this work used whole-video evaluation to compare remote signal extraction with contact-based methods. However, continuous heart rate measurement may provide more reliable insights, especially when considering interbeat interval features where accurate detection is critical.

While this study presents the first approach to remote emotion estimation using three different modalities (rPPG, rRSP, facial expressions), EDA was not incorporated, as it remains largely unexplored in remote sensing. Recent studies \cite{bhamborae2020towards, braun2024sympcam} have demonstrated the feasibility of recovering EDA signals through camera-based sensing. The integration of remote EDA into multi-modal emotion recognition frameworks could significantly enhance the estimation of sympathetic responses alongside modalities like rPPG or rRSP. In this sense, CAST-Phys can be a valuable dataset for such research lines, both in terms of providing training data and serving as a public benchmark. 

\section{Conclusions}

In this paper, we introduce the CAST-Phys dataset, a novel and realistic multi-modal emotion recognition database designed to support research on contactless physiological signal sensing for emotion analysis. The dataset comprises high-quality facial videos and physiological signals from 61 participants, with emotions elicited through stimulus videos. Our analysis highlights the important role of physiological cues in accurately estimating emotions, particularly in realistic and spontaneous scenarios where facial expressions alone may not convey sufficient information. Consistent with previous studies, our findings indicate that combining multiple modalities significantly improves the detection of emotional responses. Moreover, we propose a remote multi-modal emotion recognition baseline framework that extracts physiological signals, including remote photoplethysmography and respiration signals, from facial videos. Our results reveal a comparable performance between contact and remote multi-modal emotion recognition approaches, demonstrating the promising potential of remote physiology sensing in the field of emotion recognition.
\section{Acknowledgments}
This work is partly supported by the eSCANFace project (PID2020-114083GB-I00) funded by the Spanish Ministry of Science and Innovation.

\bibliographystyle{IEEEtran}
\bibliography{References}

@string{TAC = {IEEE Trans. Affect. Comput.}}

@String(IJCV = {Int. J. Comput. Vis.})

@string{TBE = {IEEE Trans. Biomed. Eng.}}

@String(CVPR= {IEEE Conf. Comput. Vis. Pattern Recog.})

@String(ICCV= {Int. Conf. Comput. Vis.})

@String(ECCV= {Eur. Conf. Comput. Vis.})

@string{BOE = {Biomed. Opt. Express}}

@String(BMVC= {Brit. Mach. Vis. Conf.})

@String(TIP  = {IEEE Trans. Image Process.})

@string{FG = {FG}}

@String(CVPRW= {IEEE Conf. Comput. Vis. Pattern Recog. Worksh.})

@string{JBHI = {IEEE J.Biomed.Health Inform.}}

@String(IJCV  = {IJCV})

@string{CHIL = {CHIL}}

@String(CVPR  = {CVPR})

@String(ICCV  = {ICCV})

@String(ECCV  = {ECCV})

@String(BMVC  =	{BMVC})

@String(TIP   = {IEEE TIP})

@String(CVPRW= {CVPRW})

@string{BHI ={EMBS Int. Conf. Biomed. Health Inform. BHI} }

@string{WACV = {IEEE Winter Conf. Appl. Comput. Vis.}}

@article{paltoglou2012seeing,
  title={Seeing stars of valence and arousal in blog posts},
  author={Paltoglou, Georgios and Thelwall, Michael},
  journal={IEEE Transactions on Affective Computing},
  volume={4},
  number={1},
  pages={116--123},
  year={2012},
  publisher={IEEE}
}

@article{shui2021dataset,
  title={A dataset of daily ambulatory psychological and physiological recording for emotion research},
  author={Shui, Xinyu and Zhang, Mi and Li, Zhuoran and Hu, Xin and Wang, Fei and Zhang, Dan},
  journal={Scientific data},
  volume={8},
  number={1},
  pages={161},
  year={2021},
  publisher={Nature Publishing Group UK London}
}

@article{comas2025pulseformer,
  title={PulseFormer: Continuous Remote Heart Rate Measurement Through Zoomed Time-Spectral Attention},
  author={Comas, Joaquim and Ruiz, Adri{\`a} and Sukno, Federico},
  journal={IEEE Transactions on Biometrics, Behavior, and Identity Science},
  year={2025},
  publisher={IEEE}
}

@article{Hori1992heart,
  title={Heart rate variability: frequency domain analysis},
  author={{\H{O}}ri, Zsolt and Monir, George and Weiss, Jerry and Sayhouni, Xavier and Singer, Donald H},
  journal={Cardiology clinics},
  volume={10},
  number={3},
  pages={499--533},
  year={1992},
  publisher={Elsevier}
}

@inproceedings{bhamborae2020towards,
  title={Towards contactless estimation of electrodermal activity correlates},
  author={Bhamborae, Mayur J and Flotho, Philipp and Mai, Adrian and Schneider, Elena N and Francis, Alexander L and Strauss, Daniel J},
  booktitle={2020 42nd Annual International Conference of the IEEE Engineering in Medicine \& Biology Society (EMBC)},
  pages={1799--1802},
  year={2020},
  organization={IEEE}
}

@article{braun2024sympcam,
  title={SympCam: Remote Optical Measurement of Sympathetic Arousal},
  author={Braun, Bj{\"o}rn and McDuff, Daniel and Baltrusaitis, Tadas and Streli, Paul and Moebus, Max and Holz, Christian},
  journal={arXiv preprint arXiv:2410.20552},
  year={2024}
}

@article{cheong2023py,
  title={Py-feat: Python facial expression analysis toolbox},
  author={Cheong, Jin Hyun and Jolly, Eshin and Xie, Tiankang and Byrne, Sophie and Kenney, Matthew and Chang, Luke J},
  journal={Affective Science},
  volume={4},
  number={4},
  pages={781--796},
  year={2023},
  publisher={Springer}
}

@article{Makowski2021neurokit,
     author = {Dominique Makowski and Tam Pham and Zen J. Lau and Jan C. Brammer and Fran{\c{c}}ois Lespinasse and Hung Pham and Christopher Schölzel and S. H. Annabel Chen},
     title = {{NeuroKit}2: A Python toolbox for neurophysiological signal processing},
     journal = {Behavior Research Methods},
     volume = {53},
     number = {4},
     pages = {1689--1696},
     publisher = {Springer Science and Business Media {LLC}},
     year = 2021,
     month = {feb}
 }

@article{sayette2001psychometric,
  title={A psychometric evaluation of the facial action coding system for assessing spontaneous expression},
  author={Sayette, Michael A and Cohn, Jeffrey F and Wertz, Joan M and Perrott, Michael A and Parrott, Dominic J},
  journal={Journal of nonverbal behavior},
  volume={25},
  pages={167--185},
  year={2001},
  publisher={Springer}
}

@article{ekman1978facial,
  title={Facial action coding system},
  author={Ekman, Paul and Friesen, Wallace V},
  journal={Environmental Psychology \& Nonverbal Behavior},
  year={1978}
}

@article{tao2024facial,
  title={Facial video-based non-contact emotion recognition: A multi-view features expression and fusion method},
  author={Tao, Xue and Su, Liwei and Rao, Zhi and Li, Ye and Wu, Dan and Ji, Xiaoqiang and Liu, Jikui},
  journal={Biomedical Signal Processing and Control},
  volume={96},
  pages={106608},
  year={2024},
  publisher={Elsevier}
}

@article{bradley1994measuring,
  title={Measuring emotion: the self-assessment manikin and the semantic differential},
  author={Bradley, Margaret M and Lang, Peter J},
  journal={Journal of behavior therapy and experimental psychiatry},
  volume={25},
  number={1},
  pages={49--59},
  year={1994},
  publisher={Elsevier}
}

@inproceedings{ringeval2013introducing,
  title={Introducing the RECOLA multimodal corpus of remote collaborative and affective interactions},
  author={Ringeval, Fabien and Sonderegger, Andreas and Sauer, Juergen and Lalanne, Denis},
  booktitle={2013 10th IEEE international conference and workshops on automatic face and gesture recognition (FG)},
  pages={1--8},
  year={2013},
  organization={IEEE}
}

@article{hadar2017implicit,
  title={Implicit Media Tagging and Affect Prediction from video of spontaneous facial expressions, recorded with depth camera},
  author={Hadar, Daniel},
  journal={arXiv preprint arXiv:1701.05248},
  year={2017}
}

@article{lang1997international,
  title={International affective picture system (IAPS): Technical manual and affective ratings},
  author={Lang, Peter J and Bradley, Margaret M and Cuthbert, Bruce N and others},
  journal={NIMH Center for the Study of Emotion and Attention},
  volume={1},
  number={39-58},
  pages={3},
  year={1997},
  publisher={Florida, FL}
}

@article{nardelli2015recognizing,
  title={Recognizing emotions induced by affective sounds through heart rate variability},
  author={Nardelli, Mimma and Valenza, Gaetano and Greco, Alberto and Lanata, Antonio and Scilingo, Enzo Pasquale},
  journal={IEEE Transactions on Affective Computing},
  volume={6},
  number={4},
  pages={385--394},
  year={2015},
  publisher={IEEE}
}

@article{somarathna2022virtual,
  title={Virtual reality for emotion elicitation--a review},
  author={Somarathna, Rukshani and Bednarz, Tomasz and Mohammadi, Gelareh},
  journal={IEEE Transactions on Affective Computing},
  year={2022},
  publisher={IEEE}
}

@article{russell1980circumplex,
  title={A circumplex model of affect.},
  author={Russell, James A},
  journal={Journal of personality and social psychology},
  volume={39},
  number={6},
  pages={1161},
  year={1980},
  publisher={American Psychological Association}
}

@article{peirce2019psychopy2,
  title={PsychoPy2: Experiments in behavior made easy},
  author={Peirce, Jonathan and Gray, Jeremy R and Simpson, Sol and MacAskill, Michael and H{\"o}chenberger, Richard and Sogo, Hiroyuki and Kastman, Erik and Lindel{\o}v, Jonas Kristoffer},
  journal={Behavior research methods},
  volume={51},
  pages={195--203},
  year={2019},
  publisher={Springer}
}

@article{nasoz2004emotion,
  title={Emotion recognition from physiological signals using wireless sensors for presence technologies},
  author={Nasoz, Fatma and Alvarez, Kaye and Lisetti, Christine L and Finkelstein, Neal},
  journal={Cognition, Technology \& Work},
  volume={6},
  pages={4--14},
  year={2004},
  publisher={Springer}
}

@incollection{rodriguez2021affective,
  title={Affective state-based framework for e-learning systems},
  author={Rodr{\'\i}guez, Juan Antonio and Comas, Joaquim and Binefa, Xavier},
  booktitle={Artificial Intelligence Research and Development},
  pages={357--366},
  year={2021},
  publisher={IOS Press}
}

@article{shen2009affective,
  title={Affective e-learning: Using “emotional” data to improve learning in pervasive learning environment},
  author={Shen, Liping and Wang, Minjuan and Shen, Ruimin},
  journal={Journal of Educational Technology \& Society},
  volume={12},
  number={2},
  pages={176--189},
  year={2009},
  publisher={JSTOR}
}

@article{mejbri2022trends,
  title={Trends in the use of affective computing in e-learning environments},
  author={Mejbri, Nesreen and Essalmi, Fathi and Jemni, Mohamed and Alyoubi, Bader A},
  journal={Education and Information Technologies},
  pages={1--23},
  year={2022},
  publisher={Springer}
}

@book{picard2000affective,
  title={Affective computing},
  author={Picard, Rosalind W},
  year={2000},
  publisher={MIT press}
}

@article{Liu,
author = {Liu, C. and Conn, K. and Sarkar, N. and Stone, W.},
year = {2008},
month = {09},
pages = {883 - 896},
title = {Online Affect Detection and Robot Behavior Adaptation for Intervention of Children With Autism},
volume = {24},
journal = {IEEE T Robot},
doi = {10.1109/TRO.2008.2001362}
}

@article{altameem2020facial,
  title={Facial expression recognition using human machine interaction and multi-modal visualization analysis for healthcare applications},
  author={Altameem, Torki and Altameem, Ayman},
  journal={Image and Vision Computing},
  volume={103},
  pages={104044},
  year={2020},
  publisher={Elsevier}
}

@article{chen2016facial,
  title={Facial expression recognition in video with multiple feature fusion},
  author={Chen, Junkai and Chen, Zenghai and Chi, Zheru and Fu, Hong},
  journal={IEEE Transactions on Affective Computing},
  volume={9},
  number={1},
  pages={38--50},
  year={2016},
  publisher={IEEE}
}

@inproceedings{bargal2016emotion,
  title={Emotion recognition in the wild from videos using images},
  author={Bargal, Sarah Adel and Barsoum, Emad and Ferrer, Cristian Canton and Zhang, Cha},
  booktitle={Proceedings of the 18th ACM international conference on multimodal interaction},
  pages={433--436},
  year={2016}
}

@article{tian2001recognizing,
  title={Recognizing action units for facial expression analysis},
  author={Tian, Y-I and Kanade, Takeo and Cohn, Jeffrey F},
  journal={IEEE Transactions on pattern analysis and machine intelligence},
  volume={23},
  number={2},
  pages={97--115},
  year={2001},
  publisher={IEEE}
}

@article{miranda2018amigos,
  title={Amigos: A dataset for affect, personality and mood research on individuals and groups},
  author={Miranda-Correa, Juan Abdon and Abadi, Mojtaba Khomami and Sebe, Nicu and Patras, Ioannis},
  journal={IEEE transactions on affective computing},
  volume={12},
  number={2},
  pages={479--493},
  year={2018},
  publisher={IEEE}
}

@inproceedings{comas2025beatformer,
  title={BeatFormer: Efficient motion-robust remote heart rate estimation through unsupervised spectral zoomed attention filters},
  author={Comas, Joaquim  and Sukno, Federico},
  booktitle={IEEE/CVF International Conference on Computer Vision},
  pages={520--530},
  year={2025}
}

@article{subramanian2016ascertain,
  title={ASCERTAIN: Emotion and personality recognition using commercial sensors},
  author={Subramanian, Ramanathan and Wache, Julia and Abadi, Mojtaba Khomami and Vieriu, Radu L and Winkler, Stefan and Sebe, Nicu},
  journal={IEEE Transactions on Affective Computing},
  volume={9},
  number={2},
  pages={147--160},
  year={2016},
  publisher={IEEE}
}

@article{abadi2015decaf,
  title={DECAF: MEG-based multimodal database for decoding affective physiological responses},
  author={Abadi, Mojtaba Khomami and Subramanian, Ramanathan and Kia, Seyed Mostafa and Avesani, Paolo and Patras, Ioannis and Sebe, Nicu},
  journal={IEEE Transactions on Affective Computing},
  volume={6},
  number={3},
  pages={209--222},
  year={2015},
  publisher={IEEE}
}

@article{song2019mped,
  title={MPED: A multi-modal physiological emotion database for discrete emotion recognition},
  author={Song, Tengfei and Zheng, Wenming and Lu, Cheng and Zong, Yuan and Zhang, Xilei and Cui, Zhen},
  journal={IEEE Access},
  volume={7},
  pages={12177--12191},
  year={2019},
  publisher={IEEE}
}

@article{koelstra2011deap,
  title={Deap: A database for emotion analysis; using physiological signals},
  author={Koelstra, Sander and Muhl, Christian and Soleymani, Mohammad and Lee, Jong-Seok and Yazdani, Ashkan and Ebrahimi, Touradj and Pun, Thierry and Nijholt, Anton and Patras, Ioannis},
  journal={IEEE transactions on affective computing},
  volume={3},
  number={1},
  pages={18--31},
  year={2011},
  publisher={IEEE}
}

@inproceedings{sabour2019emotional,
  title={Emotional state classification using pulse rate variability},
  author={Sabour, R Meziati and Benezeth, Yannick and Marzani, F and Nakamura, K and Gomez, R and Yang, F},
  booktitle={2019 IEEE 4th International Conference on Signal and Image Processing (ICSIP)},
  pages={86--90},
  year={2019},
  organization={IEEE}
}

@article{kreibig2010autonomic,
  title={Autonomic nervous system activity in emotion: A review},
  author={Kreibig, Sylvia D},
  journal={Biological psychology},
  volume={84},
  number={3},
  pages={394--421},
  year={2010},
  publisher={Elsevier}
}

@article{levenson2014autonomic,
  title={The autonomic nervous system and emotion},
  author={Levenson, Robert W},
  journal={Emotion review},
  volume={6},
  number={2},
  pages={100--112},
  year={2014},
  publisher={Sage Publications Sage UK: London, England}
}

@inproceedings{ouzar2022video,
  title={Video-based multimodal spontaneous emotion recognition using facial expressions and physiological signals},
  author={Ouzar, Yassine and Bousefsaf, Fr{\'e}d{\'e}ric and Djeldjli, Djamaleddine and Maaoui, Choubeila},
  booktitle={Proceedings of the IEEE/CVF conference on computer vision and pattern recognition},
  pages={2460--2469},
  year={2022}
}

@article{li2024end,
  title={End-to-end multimodal emotion recognition based on facial expressions and remote photoplethysmography signals},
  author={Li, Jixiang and Peng, Jianxin},
  journal={IEEE Journal of Biomedical and Health Informatics},
  year={2024},
  publisher={IEEE}
}

@article{comas2024physflow,
  title={PhysFlow: Skin tone transfer for remote heart rate estimation through conditional normalizing flows},
  author={Comas, Joaquim and Alomar, Antonia and Ruiz, Adria and Sukno, Federico},
  journal={arXiv preprint arXiv:2407.21519},
  year={2024}
}

@inproceedings{hsieh2022augmentation,
  title={Augmentation of rPPG benchmark datasets: Learning to remove and embed rPPG signals via double cycle consistent learning from unpaired facial videos},
  author={Hsieh, Cheng-Ju and Chung, Wei-Hao and Hsu, Chiou-Ting},
  booktitle={European Conference on Computer Vision},
  pages={372--387},
  year={2022},
  organization={Springer}
}

@inproceedings{paruchuri2024motion,
  title={Motion matters: Neural motion transfer for better camera physiological measurement},
  author={Paruchuri, Akshay and Liu, Xin and Pan, Yulu and Patel, Shwetak and McDuff, Daniel and Sengupta, Soumyadip},
  booktitle={Proceedings of the IEEE/CVF Winter Conference on Applications of Computer Vision},
  pages={5933--5942},
  year={2024}
}

@inproceedings{du2023dual,
  title={Dual-bridging with adversarial noise generation for domain adaptive rppg estimation},
  author={Du, Jingda and Liu, Si-Qi and Zhang, Bochao and Yuen, Pong C},
  booktitle={Proceedings of the IEEE/CVF Conference on Computer Vision and Pattern Recognition},
  pages={10355--10364},
  year={2023}
}

@inproceedings{lu2023neuron,
  title={Neuron structure modeling for generalizable remote physiological measurement},
  author={Lu, Hao and Yu, Zitong and Niu, Xuesong and Chen, Ying-Cong},
  booktitle={Proceedings of the IEEE/CVF conference on computer vision and pattern recognition},
  pages={18589--18599},
  year={2023}
}

@article{comas2024deep,
  title={Deep Pulse-Signal Magnification for remote Heart Rate Estimation in Compressed Videos},
  author={Comas, Joaquim and Ruiz, Adria and Sukno, Federico},
  journal={arXiv preprint arXiv:2405.02652},
  year={2024}
}

@inproceedings{revanur2021first,
  title={The first vision for vitals (v4v) challenge for non-contact video-based physiological estimation},
  author={Revanur, Ambareesh and Li, Zhihua and Ciftci, Umur A and Yin, Lijun and Jeni, L{\'a}szl{\'o} A},
  booktitle={Proceedings of the IEEE/CVF International Conference on Computer Vision},
  pages={2760--2767},
  year={2021}
}

@inproceedings{zhang2016multimodal,
  title={Multimodal spontaneous emotion corpus for human behavior analysis},
  author={Zhang, Zheng and Girard, Jeff M and Wu, Yue and Zhang, Xing and Liu, Peng and Ciftci, Umur and Canavan, Shaun and Reale, Michael and Horowitz, Andy and Yang, Huiyuan and others},
  booktitle=CVPR,
  pages={3438--3446},
  year={2016}
}

@inproceedings{gupta2023radiant,
  title={RADIANT: Better rPPG estimation using signal embeddings and Transformer},
  author={Gupta, Anup Kumar and Kumar, Rupesh and Birla, Lokendra and Gupta, Puneet},
  booktitle={Proceedings of the IEEE/CVF Winter Conference on Applications of Computer Vision},
  pages={4976--4986},
  year={2023}
}

@inproceedings{lee2020meta,
  title={Meta-rppg: Remote heart rate estimation using a transductive meta-learner},
  author={Lee, Eugene and Chen, Evan and Lee, Chen-Yi},
  booktitle=ECCV,
  pages={392--409},
  year={2020},
  organization={Springer}
}

@inproceedings{Yu2019RemotePS,
  title={Remote Photoplethysmograph Signal Measurement from Facial Videos Using Spatio-Temporal Networks},
  author={Z. Yu and Xiao-Bai Li and G. Zhao},
  booktitle=BMVC,
  year={2019}
}

@article{song2021pulsegan,
  title={PulseGAN: Learning to generate realistic pulse waveforms in remote photoplethysmography},
  author={Song,Rencheng and Chen,Huan and Cheng,Juan and Li,Chang and Liu,Yu and Chen,Xun},
  journal=JBHI,
  volume={25},
  number={5},
  pages={1373--1384},
  year={2021},
  publisher={IEEE}
}

@article{niu2019rhythmnet,
  title={Rhythmnet: End-to-end heart rate estimation from face via spatial-temporal representation},
  author={Niu, Xuesong and Shan, Shiguang and Han, Hu and Chen, Xilin},
  journal=TIP,
  volume={29},
  pages={2409--2423},
  year={2019},
  publisher={IEEE}
}

@inproceedings{yu2019remote,
  title={Remote heart rate measurement from highly compressed facial videos: an end-to-end deep learning solution with video enhancement},
  author={Yu, Zitong and Peng, Wei and Li, Xiaobai and Hong, Xiaopeng and Zhao, Guoying},
  booktitle=ICCV,
  pages={151--160},
  year={2019}
}

@inproceedings{lu2021dual,
  title={Dual-gan: Joint bvp and noise modeling for remote physiological measurement},
  author={Lu, Hao and Han, Hu and Zhou, S Kevin},
  booktitle=CVPR,
  pages={12404--12413},
  year={2021}
}

@article{poh2010non,
  title={Non-contact, automated cardiac pulse measurements using video imaging and blind source separation.},
  author={Poh, Ming-Zher and McDuff, Daniel J and Picard, Rosalind W},
  journal={Optics express},
  volume={18},
  number={10},
  pages={10762--10774},
  year={2010},
  publisher={Optical Society of America}
}

@article{jung2019utilizing,
  title={Utilizing deep learning towards multi-modal bio-sensing and vision-based affective computing},
  author={Jung, Tzyy-Ping and Sejnowski, Terrence J and others},
  journal={IEEE Transactions on Affective Computing},
  volume={13},
  number={1},
  pages={96--107},
  year={2019},
  publisher={IEEE}
}

@inproceedings{comas2020end,
  title={End-to-end facial and physiological model for affective computing and applications},
  author={Comas, Joaquim and Aspandi, Decky and Binefa, Xavier},
  booktitle={2020 15th IEEE international conference on Automatic Face and Gesture Recognition (FG 2020)},
  pages={93--100},
  year={2020},
  organization={IEEE}
}

@article{yu2023physformer++,
  title={PhysFormer++: Facial Video-Based Physiological Measurement with SlowFast Temporal Difference Transformer},
  author={Yu, Zitong and Shen, Yuming and Shi, Jingang and Zhao, Hengshuang and Cui, Yawen and Zhang, Jiehua and Torr, Philip and Zhao, Guoying},
  journal=IJCV,
  volume={131},
  number={6},
  pages={1307--1330},
  year={2023},
  publisher={Springer}
}

@inproceedings{gideon2021way,
  title={The way to my heart is through contrastive learning: Remote photoplethysmography from unlabelled video},
  author={Gideon, John and Stent, Simon},
  booktitle=ICCV,
  pages={3995--4004},
  year={2021}
}

@article{liu2020multi,
  title={Multi-task temporal shift attention networks for on-device contactless vitals measurement},
  author={Liu, Xin and Fromm, Josh and Patel, Shwetak and McDuff, Daniel},
  journal={Advances in Neural Information Processing Systems},
  volume={33},
  pages={19400--19411},
  year={2020}
}

@inproceedings{perepelkina2020hearttrack,
  title={HeartTrack: Convolutional Neural Network for Remote Video-Based Heart Rate Monitoring},
  author={Perepelkina, Olga and Artemyev, Mikhail and Churikova, Marina and Grinenko, Mikhail},
  booktitle=CVPRW,
  pages={288--289},
  year={2020}
}

@inproceedings{vspetlik2018visual,
  title={Visual heart rate estimation with convolutional neural network},
  author={{\v{S}}petl{\'\i}k, Radim and Franc, Vojtech and Matas, Jir{\'\i}},
  booktitle=BMVC,
  year={2018}
}

@inproceedings{chen2018deepphys,
  title={Deepphys: Video-based physiological measurement using convolutional attention networks},
  author={Chen, Weixuan and McDuff, Daniel},
  booktitle=ECCV,
  pages={349--365},
  year={2018}
}

@article{de2013robust,
  title={Robust pulse rate from chrominance-based rPPG},
  author={De Haan, Gerard and Jeanne, Vincent},
  journal=TBE,
  volume={60},
  number={10},
  pages={2878--2886},
  year={2013},
  publisher={IEEE}
}

@inproceedings{li2014remote,
  title={Remote heart rate measurement from face videos under realistic situations},
  author={Li, Xiaobai and Chen, Jie and Zhao, Guoying and Pietikainen, Matti},
  booktitle=CVPR,
  pages={4264--4271},
  year={2014}
}

@inproceedings{tulyakov2016self,
  title={Self-adaptive matrix completion for heart rate estimation from face videos under realistic conditions},
  author={Tulyakov, Sergey and Alameda-Pineda, Xavier and Ricci, Elisa and Yin, Lijun and Cohn, Jeffrey F and Sebe, Nicu},
  booktitle=CVPR,
  pages={2396--2404},
  year={2016}
}

@article{wang2016algorithmic,
  title={Algorithmic principles of remote PPG},
  author={Wang, Wenjin and den Brinker, Albertus C and Stuijk, Sander and De Haan, Gerard},
  journal=TBE,
  volume={64},
  number={7},
  pages={1479--1491},
  year={2016},
  publisher={IEEE}
}

@article{verkruysse2008remote,
  title={Remote plethysmographic imaging using ambient light.},
  author={Verkruysse, Wim and Svaasand, Lars O and Nelson, J Stuart},
  journal={Optics express},
  volume={16},
  number={26},
  pages={21434--21445},
  year={2008},
  publisher={Optical Society of America}
}

@article{takano2007heart,
  title={Heart rate measurement based on a time-lapse image},
  author={Takano, Chihiro and Ohta, Yuji},
  journal={Medical engineering \& physics},
  volume={29},
  number={8},
  pages={853--857},
  year={2007},
  publisher={Elsevier}
}

@article{poh2010advancements,
  title={Advancements in noncontact, multiparameter physiological measurements using a webcam},
  author={Poh, Ming-Zher and McDuff, Daniel J and Picard, Rosalind W},
  journal={IEEE transactions on biomedical engineering},
  volume={58},
  number={1},
  pages={7--11},
  year={2010},
  publisher={IEEE}
}

@inproceedings{benezeth2018remote,
  title={Remote heart rate variability for emotional state monitoring},
  author={Benezeth, Yannick and Li, Peixi and Macwan, Richard and Nakamura, Keisuke and Gomez, Randy and Yang, Fan},
  booktitle=BHI,
  pages={153--156},
  year={2018},
  organization={IEEE}
}

@article{soleymani2011multimodal,
  title={A multimodal database for affect recognition and implicit tagging},
  author={Soleymani, Mohammad and Lichtenauer, Jeroen and Pun, Thierry and Pantic, Maja},
  journal=TAC,
  volume={3},
  number={1},
  pages={42--55},
  year={2011},
  publisher={IEEE}
}

@article{sabour2021ubfc,
  title={Ubfc-phys: A multimodal database for psychophysiological studies of social stress},
  author={Sabour, Rita Meziati and Benezeth, Yannick and De Oliveira, Pierre and Chappe, Julien and Yang, Fan},
  journal=TAC,
  year={2021},
  publisher={IEEE}
}

@inproceedings{liu2021metaphys,
  title={MetaPhys: few-shot adaptation for non-contact physiological measurement},
  author={Liu, Xin and Jiang, Ziheng and Fromm, Josh and Xu, Xuhai and Patel, Shwetak and McDuff, Daniel},
  booktitle=CHIL,
  pages={154--163},
  year={2021}
}

@inproceedings{liu2023efficientphys,
  title={EfficientPhys: Enabling Simple, Fast and Accurate Camera-Based Cardiac Measurement},
  author={Liu, Xin and Hill, Brian and Jiang, Ziheng and Patel, Shwetak and McDuff, Daniel},
  booktitle=WACV,
  pages={5008--5017},
  year={2023}
}

@article{nowara2021systematic,
  title={Systematic analysis of video-based pulse measurement from compressed videos},
  author={Nowara, Ewa M and McDuff, Daniel and Veeraraghavan, Ashok},
  journal=BOE,
  volume={12},
  number={1},
  pages={494--508},
  year={2021},
  publisher={Optica Publishing Group}
}

@inproceedings{sun2022contrast,
  title={Contrast-phys: Unsupervised video-based remote physiological measurement via spatiotemporal contrast},
  author={Sun, Zhaodong and Li, Xiaobai},
  booktitle={European Conference on Computer Vision},
  pages={492--510},
  year={2022},
  organization={Springer}
}

@article{liu2024rppg,
  title={rPPG-MAE: Self-supervised pretraining with masked autoencoders for remote physiological measurements},
  author={Liu, Xin and Zhang, Yuting and Yu, Zitong and Lu, Hao and Yue, Huanjing and Yang, Jingyu},
  journal={IEEE Transactions on Multimedia},
  year={2024},
  publisher={IEEE}
}

@inproceedings{chari2024implicit,
  title={Implicit Neural Models to Extract Heart Rate from Video},
  author={Chari, Pradyumna and Harish, Anirudh Bindiganavale and Armouti, Adnan and Vilesov, Alexander and Sarda, Sanjit and Jalilian, Laleh and Kadambi, Achuta},
  booktitle={European conference on computer vision. Springer},
  year={2024}
}

@inproceedings{mcduff2017impact,
  title={The impact of video compression on remote cardiac pulse measurement using imaging photoplethysmography},
  author={McDuff, Daniel J and Blackford, Ethan B and Estepp, Justin R},
  booktitle=FG,
  pages={63--70},
  year={2017},
  organization={IEEE}
}

@inproceedings{comas2022efficient,
  title={Efficient Remote Photoplethysmography with Temporal Derivative Modules and Time-Shift Invariant Loss},
  author={Comas, Joaquim and Ruiz, Adria and Sukno, Federico},
  booktitle=CVPR,
  pages={2182--2191},
  year={2022}
}
\end{document}